\documentclass[runningheads]{llncs}

\usepackage{eccv}

\usepackage{eccvabbrv}

\usepackage{graphicx}
\usepackage{booktabs}

\usepackage[pagebackref,breaklinks,colorlinks,citecolor=eccvblue]{hyperref}

\usepackage{orcidlink}

\usepackage{epsfig}
\usepackage{graphicx}
\usepackage{amsmath}
\usepackage{amssymb}
\usepackage{times}
\usepackage{multirow}
\usepackage{pifont}
\usepackage{booktabs}
\usepackage{booktabs,multirow}
\usepackage{graphics}
\usepackage{threeparttable}
\usepackage{color}
\usepackage[normalem]{ulem}
\usepackage{float}
\usepackage{amsfonts}
\usepackage{bm}
\usepackage{enumitem}
\usepackage{array}
\usepackage{colortbl}
\definecolor{myy}{RGB}{126,95,0}
\definecolor{mygray}{gray}{.9}
\definecolor{bblue}{RGB}{30,80,120}
\definecolor{mygray1}{gray}{.7}
\usepackage{bm}
\usepackage{mathtools}
\usepackage{subcaption}
\usepackage{siunitx}
\usepackage[nopar]{lipsum}
\usepackage[export]{adjustbox}

\usepackage{url}
\usepackage{nicefrac}       
\usepackage{microtype}      
\usepackage{algorithm}
\usepackage{algorithmic}
\usepackage{wrapfig}
\usepackage{lipsum}
\usepackage{arydshln}
\usepackage{dblfloatfix}

\usepackage{soul}
\soulregister\cite7
\soulregister\ref7
\soulregister\pageref7

\newcolumntype{I}{!{\vrule width 1pt}}

\definecolor{ggray}{RGB}{127,127,127}
\definecolor{mygreen}{RGB}{93,174,86}

\makeatletter
\newcommand{\thickhline}{%
   \noalign {\ifnum 0=`}\fi \hrule height 1pt
   \futurelet \reserved@a \@xhline
}
\makeatother

\usepackage{caption}
\newcommand\figcaption{\def\@captype{figure}\caption}
\newcommand\tabcaption{\def\@captype{table}\caption}

\usepackage{algorithm}
\usepackage{listings}
\usepackage{etoolbox}
\makeatletter
\AfterEndEnvironment{algorithm}{\let\@algcomment\relax}
\AtEndEnvironment{algorithm}{\kern2pt\hrule\relax\vskip3pt\@algcomment}
\let\@algcomment\relax
\newcommand\algcomment[1]{\def\@algcomment{\footnotesize#1}}
\renewcommand\fs@ruled{\def\@fs@cfont{\bfseries}\let\@fs@capt\floatc@ruled
  \def\@fs@pre{\hrule height.8pt depth0pt \kern2pt}%
  \def\@fs@post{}%
  \def\@fs@mid{\kern2pt\hrule\kern2pt}%
  \let\@fs@iftopcapt\iftrue}
\makeatother

\usepackage{makecell}

\begin{document}

\title{
Shape2Scene: 3D Scene Representation Learning Through Pre-training on Shape Data
} 

\titlerunning{\textit{Shape2Scene}}

\author{
Tuo Feng\inst{1}\orcidlink{0000-0001-5882-3315} \and
Wenguan Wang\inst{2}\orcidlink{0000-0002-0802-9567} \and
Ruijie Quan\inst{2}\orcidlink{0000-0003-4077-1398} \and
Yi Yang\inst{2}\thanks{\textit{Corresponding author}.}\orcidlink{0000-0002-0512-880X}  
}

\authorrunning{T. Feng et al.}

\institute{ReLER, AAII, University of Technology Sydney \and
ReLER, CCAI, Zhejiang University\\
\url{https://github.com/FengZicai/S2S}
}

\maketitle

\begin{abstract}
Current 3D self-supervised learning methods of 3D scenes face a \textit{data desert} issue, resulting from the time-consuming and expensive collecting process of 3D scene data. Conversely, 3D shape datasets are easier to collect. Despite this, existing pre-training strategies on shape data offer limited potential for 3D scene understanding due to significant disparities in point quantities. To tackle these challenges, we propose Shape2Scene (S2S), a novel method that learns representations of large-scale 3D scenes from 3D shape data. We first design multi-scale and high-resolution backbones for shape and scene level 3D tasks, \ie, MH-P (point-based) and MH-V (voxel-based). MH-P/V establishes direct paths to high-resolution features that capture deep semantic information across multiple scales. This pivotal nature makes them suitable for a wide range of 3D downstream tasks that tightly rely on high-resolution features. We then employ a Shape-to-Scene strategy (S2SS) to amalgamate points from various shapes, creating a random pseudo scene (comprising multiple objects) for training data, mitigating disparities between shapes and scenes. Finally, a point-point contrastive loss (PPC) is applied for the pre-training of MH-P/V. In PPC, the inherent correspondence (\ie, point pairs) is naturally obtained in S2SS. Extensive experiments have demonstrated the transferability of 3D representations learned by MH-P/V across shape-level and scene-level 3D tasks. MH-P achieves notable performance on well-known point cloud datasets (93.8\% OA on ScanObjectNN and 87.6\% instance mIoU on ShapeNetPart). MH-V also achieves promising performance in 3D semantic segmentation and 3D object detection.
\end{abstract}

\section{Introduction}
\label{sec:introduction}
Self-supervised learning (SSL), a technique for deriving representations from unannotated data, has showcased remarkable achievements across a spectrum of domains, including natural language processing~\cite{radford2018improving, devlin2018bert, brown2020language, wei2022chain, ouyang2022training}, computer vision~\cite{he2020momentum, he2022masked}, and multi-modal learning~\cite{radford2021learning, rombach2022high, alayrac2022flamingo}. The efficacy of these techniques often hinges on extensive training with sizable datasets. However, in comparison to images and text, the \textit{data desert} issue~\cite{dong2022autoencoders} in 3D data has constrained the development of 3D SSL. Amassing extensive scene-level 3D data on a large scale demands dedicated 3D scanning equipment and platforms~\cite{geiger2013vision,dai2017scannet,armeni20163d}, resulting in financially burdensome and time-intensive data acquisition. Fortunately, shape-level datasets~\cite{wu20153d,uy2019revisiting} that encompass geometric information of individual objects are more readily accessible. Furthermore, due to the availability of open-source resources~\cite{github,Thingiverse,Sketchfab,Polycam,Smithsonian3DDigitization} and the advancement of Image-to-3D technology~\cite{threestudio2023,wu2023multiview}, obtaining such 3D data has become easier. The number of 3D shapes available online is increasingly approaching that of 2D images in the large-scale 2D dataset. This shows potential for forthcoming large-scale shape-level datasets.

\begin{figure}[t]
  \centering
      \includegraphics[width=1.01 \linewidth]{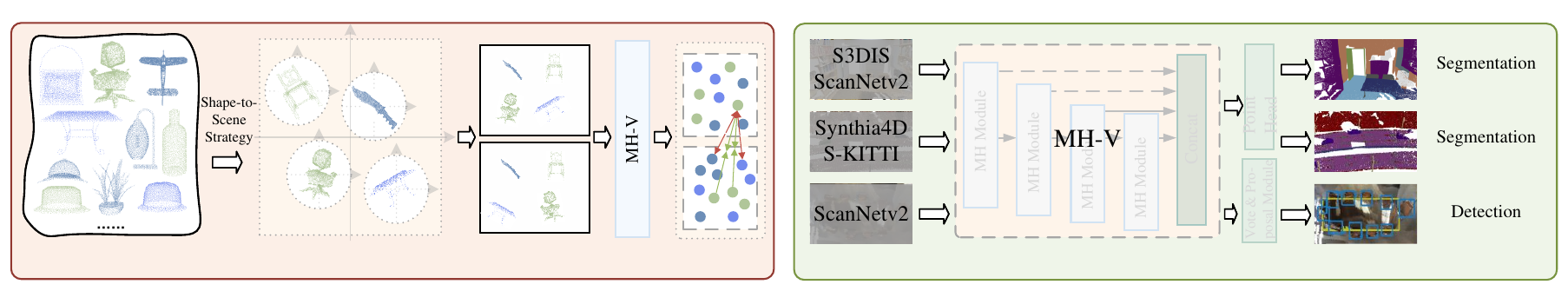}
      \put(-338,5){\scalebox{.70}{(a) Shape-to-Scene Process and Pre-Training on Shape Data}}
      \put(-153,5){\scalebox{.70}{(b) Scene-Level Downstream Tasks for Fine-Tuning}}
    \put(-243,36){\scalebox{.65}{$\mathcal{X}^1$}}
    \put(-243,14){\scalebox{.65}{$\mathcal{X}^2$}}
    \put(-30.5,45){\scalebox{.55}{\scriptsize{(\textcolor{bblue}{$+4.4\%$}) mIoU}}}
    \put(-30.5,28){\scalebox{.55}{\scriptsize{(\textcolor{bblue}{$+4.4\%$}) mIoU}}}
    \put(-30,11.5){\scalebox{.55}{\scriptsize{(\textcolor{bblue}{$+4.1\%$}) mAP}}}
\captionsetup{font=small}
\caption{\small Illustration for transferring from shape data to scene-level downstream tasks, \ie, Shape2Scene. The Shape-to-Scene strategy aggregates 4 (=$M$) shapes to one pseudo scene. Each shape is resampled and rescaled to fit onto a unit sphere. \textcolor{bblue}{Blue scores} show maximum improvements relative to training-from-scratch models. (S-KITTI stands for SemanticKITTI.) }
\label{fig:fig1}
\end{figure}
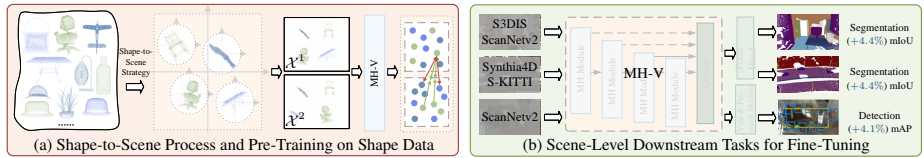

However, existing shape-level 3D SSL methods~\cite{yu2022point,liu2022masked,pang2022masked,zhang2022point,chen2023pointgpt} demonstrate limited potential for 3D scene understanding~\cite{xie2020pointcontrast} due to significant disparities in point quantities between shape and scene data. These methods prioritize the pre-training of an encoder structure to preserve global semantic representations but overlook finer high-resolution features. The decoder structure is discarded for downstream tasks. In contrast, 3D SSL methods for 3D scene understanding~\cite{xie2020pointcontrast,zhang2021self} utilize an encoder-decoder structure for high-resolution capacity. Nonetheless, these methods heavily rely on pre-training with large-scale scene-level 3D datasets, scene of which is composed of hundreds of thousands of points. Considering that 3D shape data is more readily available, the pretext task based on the shape data becomes more meaningful. The challenge lies in adapting the pre-trained model in the aforementioned setting to scene-level downstream tasks.

To tackle these challenges, we have introduced a novel 3D scene representation learning method called Shape2Scene (S2S). This method bridges the gap between shape-level and scene-level datasets from three aspects: \textbf{architecture}, \textbf{data}, and \textbf{loss}. For 3D scene understanding, including semantic segmentation and object detection~\cite{meng2021towards,meng2020weakly,yin2022semi,yin2024fusion}, high resolution capacity is an important influencing factor. However, existing methods either neglect the learning of high-resolution features or overlook the features of multi-scale structures. To tackle this, multi-scale and high-resolution \textbf{architecture} is proposed. It is principled enough to be incorporated into both point-based and voxel-based backbones, namely, MH-P and MH-V. Specifically, they map multi-scale deep semantic information to high-resolution representations, ensuring the model's transferability and generalizability for 3D scene understanding. MH-P/V incorporates a series of Multi-scale High-resolution (MH) Modules. By stacking them, MH-P/V generates high-resolution semantic representations across multiple scales, providing an advantage for high-resolution downstream tasks. In terms of \textbf{data}, we adopt a straightforward approach called the Shape-to-Scene strategy (S2SS), which involves combining points from various shapes to generate a pseudo scene with multiple objects. This strategy reduces disparities in point quantities, simulating the placement of multiple instances within a scene, and alleviates the gap between shape and scene. Regarding the \textbf{loss} function, we introduce a point-point contrastive loss (PPC) for pre-training, ensuring that MH-P/V model learns to capture intricate details and features within the scene (see Fig.\!~\ref{fig:fig1}). The inherent point pairs are naturally obtained in S2SS, making them suitable for PPC without the need for time-consuming point-level prepairing~\cite{xie2020pointcontrast, choy2019fully}.

To the best of our knowledge, S2S represents a novel method for learning 3D scene representations. Initially pre-trained on 3D shape data, it is highly adaptable to a range of downstream tasks, including large-scale 3D scene understanding. MH-P/V is a high-resolution architecture specifically designed for this method, offering three key advantages. Firstly, it provides a direct pathway to high-resolution deep semantic information across diverse scales. This fundamental characteristic makes it well-suited for various 3D downstream tasks. Secondly, it seamlessly integrates features from different scales into the point head, maintaining abstraction at varying levels and ensuring consistent semantic retention throughout the network. Features from diverse scales collaborate to collectively enhance the model's performance. Thirdly, it is versatile, as its architecture accommodates both point-based and voxel-based backbones across various 3D downstream tasks.

For a comprehensive evaluation, we examined the transferability of MH-P/V across multiple 3D tasks, yielding more promising results compared to previous 3D SSL methods. MH-P achieves an accuracy of 94.6\% on ModelNet40~\cite{wu20153d} and 93.8\% on ScanObjectNN~\cite{uy2019revisiting}. Additionally, for part segmentation, MH-P achieves an instance mIoU of 87.6\% on ShapeNetPart~\cite{yi2016scalable}. In terms of indoor semantic segmentation, MH-V achieves a mIoU of 74.1\% on S3DIS~\cite{armeni20163d} and 75.8\% on ScanNet v2~\cite{dai2017scannet}. For outdoor semantic segmentation, MH-V achieves a mIoU of 71.5\% on SemanticKITTI~\cite{behley2019semantickitti} and 84.2\% on Synthia4D~\cite{ros2016synthia}. Furthermore, MH-V demonstrates a 43.9\% mAP@0.5 in 3D object detection.

\section{Related Work}
\noindent\textbf{3D Backbones.}~3D backbone methods encompass a diverse range of techniques, including point-based methods~\cite{qi2017pointnet,qi2017pointnet++,li2018pointcnn,yang2019std,ma2021rethinking,qian2022pointnext,feng2024interpretable3d}, which directly operate on raw point cloud data. Projection-based methods~\cite{wu2018squeezeseg,zhang2020polarnet,xu2018spidercnn,tatarchenko2018tangent} transform point clouds into 2D representations. Voxel-based approaches~\cite{choy20194d,yan2018second,riegler2017octnet,graham20183d,klokov2017escape,zhu2021cylindrical} employ voxelization techniques and apply sparse convolution exclusively to non-empty voxels. The 3D backbones diverge in network structure for shape-level and scene-level tasks. Networks for shape classification commonly employ successive downsampling to acquire high-level semantic features while maintaining a lower resolution. Conversely, for extracting deeper features with higher resolution, well-known networks for scene-level tasks often incorporate architectures like U-Net or hourglass-like networks~\cite{qi2017pointnet++,li2018pointcnn,ma2021rethinking,qian2022pointnext,wu2018squeezeseg,zhang2020polarnet,choy20194d,yan2018second,riegler2017octnet,graham20183d,klokov2017escape}. For contrastive methods~\cite{afham2022crosspoint, jing2020self} and multi-modal methods~\cite{xue2023ulip, xue2023ulip2}, backbones originally designed for shape classification tasks are employed during pre-training. However, it has been demonstrated that the features extracted by these low-resolution backbones are not suitable for tasks that require high resolution representations~\cite{sun2018fishnet}. Directly utilizing high-resolution shallow features for region and point level tasks, however, does not yield satisfactory results~\cite{sun2018fishnet}.

We present a multi-scale, high-resolution mechanism that preserves high-resolution features along with high-level semantic information. This approach improves the applicability of pre-trained high-level semantic features, making them more suitable for point-level tasks. This also ensures the preservation and mutual refinement of features across a wide range of scales, each offering distinct levels of abstraction for the point cloud. Retaining all these scale-specific features enhances diversity, and their complementary nature enables effective mutual refinement.

\noindent\textbf{SSL for Point Clouds.}~SSL has garnered attention in point cloud representation learning, evident through extensive research~\cite{wang2021unsupervised, zhang2021self, yin2022proposalcontrast, xie2020pointcontrast, sauder2019self, yin2022semi, achlioptas2018learning, li2018so, yang2018foldingnet, rao2020global, eckart2021self, yan2022iae, feng2023clustering}. Among the diverse landscape of SSL frameworks tailored for point clouds, a particular subset, known as contrastive methods~\cite{xie2020pointcontrast, zhang2021self}, has gained substantial attention. Among these methods, PointContrast~\cite{xie2020pointcontrast} stands out by harnessing the power of contrastive learning, which enhances 3D representations through feature comparisons extracted from different viewpoints. DepthContrast~\cite{zhang2021self} introduces an innovative approach using augmented depth maps to facilitate instance discrimination and global feature extraction, leading to a demonstrably improved understanding of 3D point clouds.

Generative methods, particularly those involving masked point modeling, have emerged as a focal point of research in the field~\cite{pang2022masked, yu2022point, zhang2022point, min2022voxel}. OcCo~\cite{wang2021unsupervised} utilizes an encoder-decoder architecture to reconstruct occluded point clouds. Its primary goal is to restore point clouds from observations made from camera views of the occluded version. Point-BERT~\cite{yu2022point} introduces a BERT-style pre-training approach for 3D point clouds, predicting masked regions and drawing inspiration from the concepts in dVAE~\cite{rolfe2016discrete}. Notably, both Point-BERT and its counterpart Point-MAE~\cite{pang2022masked} utilize pre-training techniques inspired by BERT~\cite{devlin2018bert} and MAE~\cite{he2022masked}, consistently demonstrating impressive performance across various downstream tasks. Point-M2AE~\cite{zhang2022point} implements an MAE-style framework featuring a hierarchical transformer for multi-scale pre-training, achieving state-of-the-art results. Meanwhile, PointGPT~\cite{chen2023pointgpt} addresses challenges such as object shape leakage by leveraging auto-regressive pre-training, effectively reducing positional information leakage and enhancing the model's generalization capabilities.

Pre-training data for generative methods typically involves 3D shape data. Conversely, previous SSL methods for 3D scene understanding either utilize pre-training data derived from 3D scene data~\cite{xie2020pointcontrast,long2023pointclustering} or consider 3D scene data as an indispensable component of pre-training~\cite{zhang2021self}. In the realm of 3D scene understanding, our S2S method fundamentally distinguishes itself from various aforementioned methods. MH-VH, pre-trained on shapes, outperforms various methods \cite{xie2020pointcontrast,long2023pointclustering,zhang2021self}, which are pre-trained on scenes. Despite attempts by 4DContrast~\cite{chen20224dcontrast} to generate pseudo-scenes by aggregating objects with ScanNet (scene), our S2SS fundamentally differs from it, as we aggregate shapes independently of scene data. In summary, previous studies ~\cite{xie2020pointcontrast,long2023pointclustering,zhang2021self,chen20224dcontrast} inevitably relied on substantial amounts of scene data (ScanNet v2) for pre-training.

\section{Methodology}
\label{sec:method}
This section introduces the novel 3D scene representation learning method, S2S, learning representations of large-scale 3D scenes from 3D shape data. In \S\!~\ref{sec:shape1} and \S\!~\ref{sec:shape2}, we commence with the introduction of MH-P/V and MH module. MH-P/V is a versatile point cloud pre-training backbone, providing a flexible foundation for SSL of 3D point clouds. It is applicable to a range of 3D downstream tasks, spanning from shape-level tasks to scene-level tasks, especially for tasks that focus on learning high-resolution features. Subsequently, we delve into the specifics of S2SS in \S\!~\ref{sec:scene1}, along with a detailed description of PPC in \S\!~\ref{sec:scene2}.

Suppose that the input is $\mathcal{I}\!=\!\{\mathcal{P}^k\}_{k}^K$, \ie, a set with $K$ training samples. Here $\mathcal{P}^{k\!}\!=\!\{p^k_{n\!}\!\in\!\mathbb{R}^{3+a}\}_{n\!}^{N}$ is the $k$-\textit{th} sample containing $N$~points with 3D position and other auxiliary information (\eg, color). We represent the MH-P/V backbone and point head as $\psi$ and $\theta$, respectively.

\begin{figure*}[t]
  \centering
    \includegraphics[width=1.01 \linewidth]{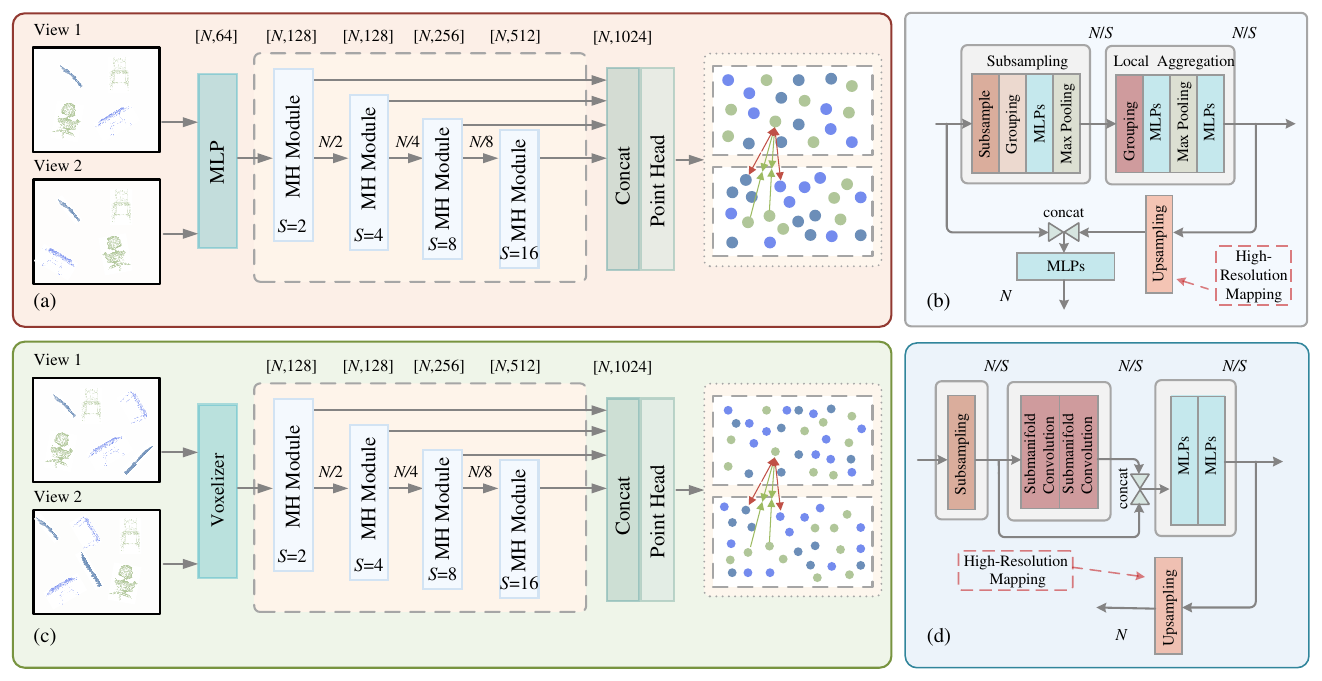}
    \put(-232,161){\scalebox{.55}{$x_2^h$}}
    \put(-221,154.5){\scalebox{.55}{$x_4^h$}}
    \put(-210,148.5){\scalebox{.55}{$x_8^h$}}
    \put(-203,141){\scalebox{.55}{$x_{16}^h$}}
    \put(-232,73){\scalebox{.55}{$x_2^h$}}
    \put(-221,66.7){\scalebox{.55}{$x_4^h$}}
    \put(-210,60.7){\scalebox{.55}{$x_8^h$}}
    \put(-203,53){\scalebox{.55}{$x_{16}^h$}}
    \put(-280,110){\scalebox{.8}{$\psi$}}
    \put(-280,22){\scalebox{.8}{$\psi$}}
    \put(-179,145.5){\scalebox{.8}{\rotatebox{90}{$\theta$}}}
    \put(-179,59){\scalebox{.8}{\rotatebox{90}{$\theta$}}}
    \put(-327,171){\scalebox{.55}{$\mathcal{T}(\mathcal{P})$}}
    \put(-327,135){\scalebox{.55}{$\mathcal{T}^{*}(\mathcal{P})$}}
    \put(-327,83.5){\scalebox{.55}{$\mathcal{T}(\mathcal{P})$}}
    \put(-327,47.5){\scalebox{.55}{$\mathcal{T}^{*}(\mathcal{P})$}}
    \put(-66,100){\scalebox{.65}{$x_{s}^h$}}
    \put(-20,150){\scalebox{.65}{$x_{s}$}}
    \put(-57,21.5){\scalebox{.65}{$x_{s}^h$}}
    \put(-20,60){\scalebox{.65}{$x_{s}$}}
\captionsetup{font=small}
\caption{\small (a) The overview of the MH-P backbone during pre-training, with contrastive loss (Eq.~(\ref{eq:PointInfoNCE})). (b) The Multi-scale High-resolution (MH) Module of MH-P (see \S\!~\ref{sec:shape1} for details.) (c) The overview of the MH-V backbone during pre-training, with contrastive loss (Eq.~(\ref{eq:PointInfoNCE})). (d) The Multi-scale High-resolution (MH) Module of MH-V (see \S\!~\ref{sec:shape2} for details).}
\label{fig:framework}
\end{figure*}
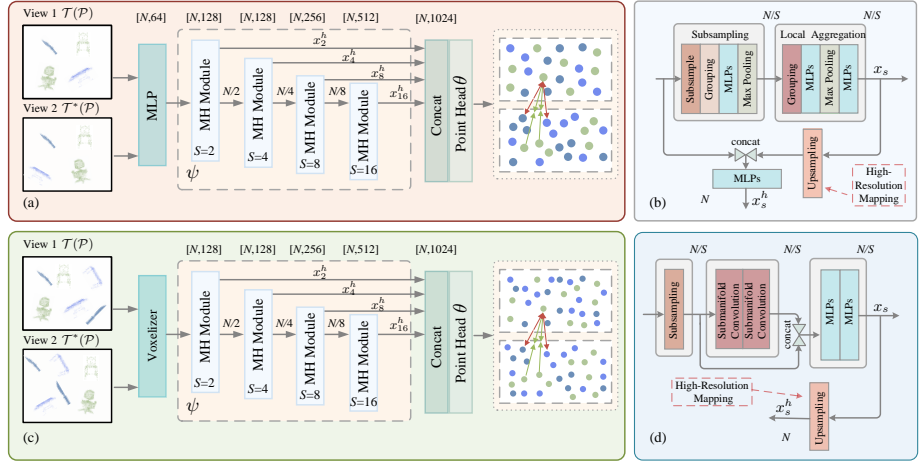

\subsection{Point-based MH Backbone}
\label{sec:shape1}
Fig.\!~\ref{fig:framework} (a) provides an overview of MH-P designed for the shape-level tasks. MH-P comprises MH modules. High-resolution features are learned by point head $\theta$.

\noindent\textbf{MH Module of MH-P.}~As shown in Fig.\!~\ref{fig:framework} (b), each MH module is divided into three parts: subsampling, local aggregation, and high-resolution mapping. In each MH module, we adopt a set abstraction (SA) block~\cite{qi2017pointnet++} for subsampling. One SA block comprises several key components: a subsampling layer that reduces the resolution of incoming points, a grouping layer responsible for identifying neighbors for each point, a series of shared multi-layer perceptrons (MLPs) designed for feature extraction, and a max pooling layer that combines features from neighboring points. Local aggregation consists of a grouping layer, MLPs, and a max pooling layer. High-resolution mapping employs nearest neighbor interpolation to assign low-resolution high-dimensional features to the nearest high-resolution points.

\noindent\textbf{MH-P.}~Assuming one MH module with the scale of $S$, both spatial geometry and semantic information are learned through subsampling and local aggregation. We denote the learned deep semantic representation as the semantic feature $x_s$ for $S$. After undergoing high-resolution mapping, $x_s$ is ultimately transformed into a high-resolution format, yielding a high-resolution semantic representation $x_s^h$. It will be used as the input for the next scale. As the scale increases, it progressively gathers more sophisticated semantic information from lower-resolution data. Stacked modules generate point-level semantic representations at multiple scales, which are advantageous for high-resolution tasks. The overall network structure is shown in Fig.\!~\ref{fig:framework} (a).

For shape-level high-resolution tasks, \eg, part segmentation, $\{x_s^h\}_s$ are integrated as the input of $\theta$. This ensures the consistent retention of semantics throughout the network, and the collective synergy of features from various scales enhances the model's performance. MH-P establishes a direct pathway to high-resolution features that capture rich semantic information across multiple scales. By translating multi-scale deep semantic information into high-resolution/point-level representations, MH-P ensures that features originally trained on 3D shape data can seamlessly adapt to a variety of 3D downstream tasks, especially for tasks that necessitate high-resolution representations. This pivotal characteristic of MH-P makes it exceptionally well-suited for high-resolution representations in part segmentation.

The shape classification task does not emphasize high-resolution representations but instead extracts global features. To adapt to this requirement, we only need to make minor modifications to MH-P. In each MH module, $x_s$ passes through one more max-pooling layer, obtaining $x_s^g$, which represents the global feature of scale $S$. Subsequently, we seamlessly integrate ${x_s^g}$ from different scales into the input of one shape head, preserving abstractions at various levels. The surprising thing is that, during the transition from high-resolution representations to global features, MH-P still achieves better performance on downstream tasks (see \S\!~\ref{sec:downstream}). We attribute this to the network's capacity to acquire essential abstractions at various scales. This guarantees the preservation of vital information and its effective utilization across different scales, thereby maintaining performance even with the shifts towards global features.

\subsection{Voxel-based MH Backbone}
\label{sec:shape2}

There are significant differences in dataset statistics between shape-level and scene-level data, with $\sim$1k input points for ModelNet40~\cite{wu20153d}, $\sim$40k input points for ScanNet v2~\cite{dai2017scannet}, and $\sim$956K input points for S3DIS~\cite{armeni20163d}. Following the methods for processing large-scale point clouds~\cite{choy20194d,zhu2021cylindrical}, we design a voxel-based MH backbone with MH modules. Fig.\!~\ref{fig:framework} (c) provides an overview of MH-V, and Fig.\!~\ref{fig:framework} (d) shows the MH module for MH-V. Different from hourglass-like networks, MH-V follows a similar concept to MH-P and learns high-resolution features at the scene level across multiple scales. For simplicity, we retain the symbols used in \S\!~\ref{sec:shape1}.

\noindent\textbf{MH Module of MH-V.}~As depicted in Fig.~\ref{fig:framework} (d), this MH module possesses a structure similar to that of the MH module of MH-P (in \S\!~\ref{sec:shape1}), including featuring subsampling, local aggregator, and upsampling functionalities. Specifically, the voxelizer first defines indices and inverse indices between points and voxels at different scales, as well as indices and inverse indices between voxels at different scales. These indices will be applied for subsampling, \ie, point features with the same voxel index will be averaged and then treated as features of the corresponding voxel. The inverse indices will be used for upsampling or high-resolution mapping, \ie, voxel features will be assigned to corresponding points according to the inverse indices. The local aggregator employs stacked submanifold convolutions to extract local features, combines them with features from skip connections, and then processes them through a series of MLPs to derive a deep semantic representation $x_s$. While upsampling, $x_s$ is ultimately transformed into a high-resolution/point-level representation $x_s^h$.

\noindent\textbf{MH-V.}~In MH-V, $x_s$ of the MH module for $S$ is taken as input of the MH module for the next scale. Similar to MH-P, we employ high-resolution mappings within each MH module, and all $\{x_s^h\}_s$ are integrated as the input of $\theta$. Our MH-V not only effectively manages the substantial volume of data at the scene level but also benefits from direct access to high-resolution data and integrated features from various scales.

Our MH-V differs fundamentally in structure from other hourglass-like voxel-based 3D networks (\eg, SparseConvNet~\cite{graham20183d} and MinkUNet~\cite{choy20194d}). MH-V serves as a backbone network aimed at pursuing multi-scale high-resolution features. The significance of high-resolution 3D network has been validated in \cite{liu2019point2sequence,tang2020searching}, thus affirming the effectiveness of our MH-V. However, unlike previous methods, we achieve downsampling and upsampling entirely through indices and inverse indices, eliminating the need for \textbf{regular} sparse convolution or trilinear interpolation of nearest neighbor voxels. Our approach is more streamlined, efficient, and not affected by the issue of blurring valuable information caused by regular sparse convolution~\cite{chen2022focal,feng2024lsk3dnet}.

\subsection{Shape-to-Scene Strategy}
\label{sec:scene1}
Previous work~\cite{xie2020pointcontrast} has demonstrated the significance of point-level representations for 3D scene understanding. Directly training on a single shape and acquiring a global representation might be inadequate for scene-level tasks. To mitigate this concern, it could be advantageous to directly pre-train the network on complex scenes containing multiple objects to more accurately align with the target distributions~\cite{xie2020pointcontrast}. Therefore, a series of SSL methods for 3D scene understanding have depended on scene data~\cite{xie2020pointcontrast,long2023pointclustering,zhang2021self,chen20224dcontrast}. However, collecting 3D scene data is financially burdensome and time-intensive. With the development of open-source platforms and Image-to-3D technology, 3D shape data has become more readily accessible. Here, we revisit the challenge of extending 3D SSL from 3D shape data to 3D scene data.

We propose S2SS to aggregate multiple objects (shapes) to generate pseudo scene-level training data. This strategy is entirely different from previous methods~\cite{xie2020pointcontrast,long2023pointclustering,zhang2021self,chen20224dcontrast} in terms of how the pre-training data is created. Moreover, it brings the additional benefit that it even allows the natural derivation of positive and negative pairs by leveraging known shape correspondences, without the need for any extra computation. As shown in Fig.~\!\ref{fig:fig1} (a), given $M$ shapes, we undergo random presampling with 2,048 points for each shape. Each shape is then normalized by rescaling it to fit onto a unit sphere. Through translation, we position various shapes within a shared world coordinate. The Euclidean distance between the barycenters of any two shapes is ensured to be greater than 2, preventing any overlap among shapes. The final output yields pseudo scene data containing $M$ shapes: $\mathcal{P}_{s}\!=\!\{p_{n}\}_{n=1}^{M \times 2048}$. Next, we use two rigid transformations, denoted as $\{\mathcal{T}(\cdot),\mathcal{T}^{*}(\cdot)\}$, randomly sampled from the transformation set $\mathcal{T}$. The transformations in $\mathcal{T}$ are applied to the pseudo scene data, and $\mathcal{T}$ encompasses rotation, translation, and scaling, \etc. We produce two views $\mathcal{X}^1\!\!=\!\!\mathcal{T}(\mathcal{P}_{s})$ and $\mathcal{X}^2\!\!=\!\!\mathcal{T}^{*}(\mathcal{P}_{s})$, which are aligned within the same world coordinates.

\subsection{Point-Point Contrastive Loss}
\label{sec:scene2}
The fundamental principle of contrastive learning hinges on the concept of \textit{invariance learning}~\cite{yvette2011noether, he2020momentum, kong2023understanding}. This entails that the abstraction of semantics generally remains either invariant or equivariant~\cite{dangovski2021equivariant} in the face of various transformed perspectives, such as augmentations~\cite{chen2020simple}. Previous work~\cite{xie2020pointcontrast, wang2021exploring} has demonstrated the significance of point-level representations over global representations. Similarly, our designed MH-P/V aims to acquire point-level/high-resolution features. Therefore, on top of the PointInfoNCE loss~\cite{xie2020pointcontrast}, we introduce PPC. However, PPC shares a close relationship with S2SS, where the inherent point pairs are naturally obtained without relying on the time-consuming point-level pairing method, FCGF~\cite{xie2020pointcontrast,choy2019fully}. During pre-training, we utilize $\theta$ to acquire high-resolution 3D representations. Given input data $p$, the model $\mathcal{E}\!=\!\phi\circ\psi$ is optimized by:
\begin{equation}\label{eq:PointInfoNCE}{\small
 {\mathcal{L}_{\text{PPC}}}\!=\!-\!\!\!\!\!\sum_{(u,v) \in \mathcal{O}_p}\!\!\!\log\!\frac{\exp\Big(\mathcal{E}\big(\mathcal{T}(p_u)\big)\!\cdot\!\mathcal{E}\big(\mathcal{T}^{*}(p_v)\big) /\tau \Big)}{\sum\nolimits_{(\cdot, w) \in \mathcal{O}_p}\!\!\exp\Big(\mathcal{E}\big(\mathcal{T}(p_u)\big)\!\cdot\!\mathcal{E}\big(\mathcal{T}^{*}(p_w)\big) /\tau \Big)},}
\end{equation}
where $\mathcal{E}\!=\!\theta\circ\psi$ represents the whole model, $\mathcal{E}\big(\mathcal{T}(p_{*})\big)$ denotes the representation learned by $\mathcal{E}$; $\tau$ is a temperature hyperparameter; and $\mathcal{O}_p\!=\!\{\!\thinspace(n,n)\thinspace|\thinspace p_n\!\in\!\mathcal{P}_{s}\} \cup \{\thinspace(i,j)\thinspace|\thinspace p_i^k\!\in\!\mathcal{P}^{k\!} \cap p_j^k\!\in\!\mathcal{P}^{k\!}\}$ denotes the collection of all positive matches derived from two different views. This approach focuses solely on points that have at least one corresponding match and regards additional non-matched points as negative matches. In a matched pair $(u, v)\!\in\!\mathcal{O}_p$, the point feature $\mathcal{E}\big(\mathcal{T}(p_u)\big)$ operates as the query, and $\mathcal{E}\big(\mathcal{T}(p_v)\big)$ functions as the positive key. We employ the point feature $\mathcal{E}\big(\mathcal{T}^{*}(p_w)\big)$, where $\exists (\cdot, w)\!\in\!\mathcal{O}_p$ and $w\!\neq\!v$, as the pool of negative keys. Following \cite{xie2020pointcontrast}, we choose a subset of 4,096 matched pairs from $\mathcal{O}_p$ for scene-level downstream tasks. However, we empirically select a subset of 2,048 matched pairs from $\mathcal{O}_p$ for shape-level downstream tasks. Additionally, we present a PyTorch-style pseudo-code for PPC in the supplementary.

The way we obtain $\mathcal{O}_p$ fundamentally differs from previous methods~\cite{xie2020pointcontrast,zhang2021self}. On the one hand, rigid transformations maintain the point cloud's order, ensuring a one-to-one correspondence (\ie, $\{\!\thinspace(n,n)\thinspace|\thinspace p_n\!\in\!\mathcal{P}_{s}\}$) between $\mathcal{X}^1$ and $\mathcal{X}^2$. On the other hand, points originating from the same shape are regarded as positive pairs (\ie, $\{\thinspace(i,j)\thinspace|\thinspace p_i^k\!\in\!\mathcal{P}^{k\!} \cap p_j^k\!\in\!\mathcal{P}^{k\!}\}$) across the two views. In this way, we create the scene-level point pair data that is original from single objects. These designs fully exploit the correspondences between points and the relationships within points in the shape.

\section{Experiment}
\label{sec:experiment}

In \S\!~\ref{sec:pretraining}, we introduce the self-supervised pre-training settings for MH-P/V. Subsequently, we present the supervised fine-tuning performance on shape-level and scene-level downstream tasks in \S\!~\ref{sec:downstream} and \S\!~\ref{sec:downstream2}, respectively. Furthermore, we conduct ablation studies in \S\!~\ref{sec:ablation} to validate the effectiveness of each part of our method.

\subsection{Pre-Training Settings}
\label{sec:pretraining}

\noindent\textbf{MH-P and Data Settings.}~We conduct pre-training of MH-PS on ShapeNet dataset~\cite{chang2015shapenet} and MH-PH on both the labeled hybrid dataset (LHD) and the unlabeled hybrid dataset (UHD)~\cite{chen2023pointgpt}. MH-PS is a backbone network composed of four MH modules. ShapeNet comprises $\sim$52K synthetic 3D shapes in 55 categories. MH-PS is trained without any post-pre-training, aligning with previous SSL methods~\cite{pang2022masked, yu2022point, zhang2022point, liu2022masked} to enable direct comparison with them. Moreover, MH-PH is pre-trained on LHD and UHD. MH-PH also consists of four MH modules. However, to explore a high capacity model, we scale it by increasing the network width and the number of local aggregation within the MH modules. The UHD, used for self-supervised pre-training, aggregates point clouds from various sources such as ShapeNet~\cite{chang2015shapenet}, S3DIS~\cite{armeni20163d} for indoor scenes, and Semantic3D~\cite{hackel2017semantic3d} for outdoor scenes, \etc. In total, UHD comprises around 300K point clouds. Conversely, LHD, utilized for post-pre-training, aligns the label semantics from diverse datasets, including ShapeNet~\cite{chang2015shapenet}, S3DIS~\cite{armeni20163d}, and other sources, encompassing 87 categories and roughly 200K point clouds in total.

\noindent\textbf{MH-V and Data Settings.}~We employ a similar setup as described above. The difference lies in combining $M$ shapes to form a scene for pre-training. We denote the MH-V pre-trained on ShapeNet as MH-VS, and the MH-V pre-trained on UHD and LHD as MH-VH. We also scale MH-VH by increasing the network width and the number of local aggregators within the MH modules.

\noindent\textbf{Pre-Training Setups.}~We extract the input shapes by sampling 2,048 points from each raw point cloud, and compose $M$ shapes to form a scene. See \S\!~\ref{sec:scene1} and \S\!~\ref{sec:ablation}. The MH-P model undergoes pre-training for 600 epochs with a batch size of 10. We utilize the AdamW optimizer~\cite{loshchilov2017decoupled} with an initial learning rate of 0.001, and a weight decay of 0.05. Additionally, we employ cosine learning rate decay~\cite{loshchilovstochastic} based on our empirical findings. The MH-V model undergoes pre-training for 800 epochs on two V100 GPUs with a batch size of 10. Other configurations align with those used for MH-P.

\subsection{Shape Level Downstream Tasks}
\label{sec:downstream}
We perform supervised fine-tuning of MH-PS and MH-PH on three classical downstream tasks. Each shape-level downstream task has thousands of input points.

\begin{table}[t]
  \begin{minipage}[t]{0.49\linewidth}
        \captionsetup{font=small}
  \caption{\small \textbf{Classification results on ScanObjectNN~\cite{uy2019revisiting} and ModelNet40~\cite{wu20153d} datasets (\S\!~\ref{sec:downstream}).} We report the overall accuracy (\%).} \label{tab:scanobjectnn}
        \centering
        \small
        \setlength\tabcolsep{0.5pt}
        \renewcommand\arraystretch{1.1}
        \resizebox{\linewidth}{!}{
        \begin{tabular}{rcccc}
        \toprule[0.95pt]
        \multicolumn{1}{c}{} & \multicolumn{3}{c}{ScanObjectNN} & \\  \cmidrule(lr){2-4} 
        \multicolumn{1}{c}{\multirow{-2}{*}{Methods}}  & OBJ\_BG & OBJ\_ONLY & PB\_T50\_RS& \multicolumn{1}{c}{\multirow{-2}{*}{ModelNet40}}\\
        \midrule[0.6pt]
        \multicolumn{5}{c}{\textit{Supervised Learning Only}}\\
        \midrule[0.6pt]
        DGCNN~\cite{phan2018dgcnn}& 82.8 & 86.2 & 78.1 &92.9\\
        PointNet++~\cite{qi2017pointnet++} & - & - & 77.9 &90.5\\
        PointMLP~\cite{ma2021rethinking} & - & - & 85.4 &94.5\\
        PointNeXt~\cite{qian2022pointnext} & - & - & 87.7 & 94.0\\
        \midrule[0.6pt]
        \multicolumn{5}{c}{\textit{Pre-training on ShapeNet}}\\
        \midrule[0.6pt]
        Point-BERT~\cite{yu2022point} & 87.4 & 88.1 & 83.1&93.2 \\
        MaskPoint~\cite{liu2022masked} & 89.3 & 88.1 & 84.3& 93.8\\
        Point-MAE~\cite{pang2022masked} & 90.0 & 88.3 & 85.2& 93.8 \\
        Point-M2AE~\cite{zhang2022point} & 91.2 & 88.8 & 86.4& 94.0 \\
        PointGPT-S~\cite{chen2023pointgpt} & 91.6 & 90.0 & 86.9& 94.0 \\
        \rowcolor{mygray}
        MH-PS (Ours) & \textbf{92.4} & \textbf{90.8} & \textbf{87.8}& \textbf{94.1}\\
        \midrule[0.6pt]
        \multicolumn{5}{c}{\textit{Pre-training on UHD and LHD}}\\
        \midrule[0.6pt]
        PointGPT-B~\cite{chen2023pointgpt} & 95.8 & 95.2 & 91.9& 94.4 \\
        PointGPT-L~\cite{chen2023pointgpt} & 97.2 & 96.6 & 93.4& \textbf{94.7} \\
        \rowcolor{mygray}
        MH-PH (Ours) & \textbf{97.4} & \textbf{96.8} & \textbf{93.8} & 94.6\\
        \bottomrule[0.95pt]
        \end{tabular}}
        \end{minipage}
        \hfill
        \begin{minipage}[t]{0.49\linewidth}
        \captionsetup{font=small}
        \caption{\small
        \textbf{Segmentation results on the Shape- NetPart~\cite{yi2016scalable} dataset (\S\!~\ref{sec:downstream}).} We report the Class mIoU and Instance mIoU.}
        \label{tab:partsegmentation}
        \centering
        \small
        \setlength\tabcolsep{3.3pt}
        \renewcommand\arraystretch{1.05}
        \resizebox{\linewidth}{!}{
        \begin{tabular}{rccc}
        \toprule[0.95pt]
        \multicolumn{1}{c}{} &\multicolumn{1}{c}{} & \multicolumn{2}{c}{ShapeNetPart} \\  
        \cmidrule(lr){3-4}
        \multicolumn{1}{c}{\multirow{-2}{*}{Methods}} &\multicolumn{1}{c}{\multirow{-2}{*}{Backbone}} & Cls. mIoU(\%) & Inst. mIoU(\%) \\
        \midrule[0.6pt]
        \multicolumn{4}{c}{\textit{Supervised Learning Only}}\\
        \midrule[0.6pt]
        PointNet++~\cite{qi2017pointnet++}& PointNet++ &81.9 & 85.1\\
        DGCNN~\cite{phan2018dgcnn} & DGCNN & 82.3 & 85.2\\
        \midrule[0.6pt]
        \multicolumn{4}{c}{\textit{Pre-training on ScanNet}}\\
        \midrule[0.6pt]
        DGCNN~\cite{wang2021unsupervised} & DGCNN &-&85.0\\ 
        PointNet++~\cite{long2023pointclustering}& PointNet++ &-&85.9\\
        PointViT~\cite{long2023pointclustering}& Transformer &-&86.7\\
        \midrule[0.6pt]
        \multicolumn{4}{c}{\textit{Pre-training on ShapeNet}}\\
        \midrule[0.6pt]
        Point-BERT~\cite{yu2022point}& Transformer & 84.1 & 85.6 \\
        Point-MAE~\cite{pang2022masked} & Transformer & - & 86.1 \\
        Point-M2AE~\cite{zhang2022point}& Transformer & 84.9 & 86.5 \\
        PointGPT-S~\cite{chen2023pointgpt} & Transformer & 84.1 & 86.2 \\
        \rowcolor{mygray}
        MH-PS (Ours) & S2S & \textbf{85.3} & \textbf{87.1}\\
        \midrule[0.6pt]
        \multicolumn{4}{c}{\textit{Pre-training on UHD and LHD}}\\
        \midrule[0.6pt]
        PointGPT-B~\cite{chen2023pointgpt} & Transformer & 84.5 & 86.5 \\
        PointGPT-L~\cite{chen2023pointgpt} & Transformer & 84.8 & 86.6 \\
        \rowcolor{mygray}
        MH-PH (Ours)& S2S & \textbf{85.5} &\textbf{87.6}\\
        \bottomrule[0.95pt]
        \end{tabular}}
        \end{minipage}
\end{table}

\noindent\textbf{Shape Classification on ScanObjectNN.}~ScanObjectNN stands out as one of the most formidable 3D datasets, encompassing 15K objects extracted from real-world indoor scans~\cite{uy2019revisiting}. Within the dataset, there are three commonly utilized data splits, namely OBJ\_ONLY (object only), OBJ\_BG (with background), and the PB\_T50\_RS (with background and manual perturbations). Following~\cite{liu2022masked,pang2022masked,uy2019revisiting,yu2022point,zhang2022point,chen2023pointgpt}, we conduct experiments on the three splits mentioned above and reported overall accuracy (OA) on the respective test sets. As evidenced in Tab.\!~\ref{tab:scanobjectnn}, our approach, MH-PS, outperforms PointGPT-S on all three data splits' test sets. Our performance improvements can be attributed to the effective utilization of the multi-scale mechanism. It contributes to the preservation of global features and results in enhanced performance at the shape level tasks, yielding an improvement of approximately 0.8\% to 0.9\%. Furthermore, MH-PH showcases better performance compared to PointGPT-L.

\noindent\textbf{Shape Classification on ModelNet40.}~ModelNet40 is a classical dataset for synthetic 3D object recognition~\cite{wu20153d}. It contains $\sim$12K meshed 3D CAD objects of 40 classes. For fair comparisons, the standard voting method~\cite{liu2019relation} is used during testing. The experimental results are presented in Tab.\!~\ref{tab:scanobjectnn}, MH-P achieves comparable performance to PointGPT in two pre-training data settings. ModelNet40 has been thoroughly explored (saturated around 94\%), and this performance is already convincing enough. Moreover, MH-PS surpasses other MAE-style and BERT-style pre-training frameworks.

\noindent\textbf{Part Segmentation.}~We evaluate MH-P on ShapeNetPart~\cite{yi2016scalable} for part segmentation task (a high-resolution task), which predicts per-point part labels for a known object. The ShapeNetPart dataset consists of 17K objects across 16 categories. The point clouds are sampled into 2,048 points. Tab.\!~\ref{tab:partsegmentation} provides the instance mean Intersection over Union (Inst. mIoU) and class mean Intersection over Union (Cls. mIoU) results. Despite the strong Transformer architectures (MAE-style and BERT-style backbones), our MH-PS even surpasses the performance of PointGPT-B and PointGPT-L, which are pre-trained on UHD and LHD. This underscores the effectiveness of our method, as MH-P captures high-resolution semantic features from deep semantic information across multiple scales. Additionally, our MH-PH achieves state-of-the-art performance on both Cls. mIoU and Inst. mIoU metrics.

\begin{table}[t]
\begin{minipage}[t]{.49\textwidth}
\captionsetup{font=small}
\caption{\small
\textbf{Segmentation results on S3DIS Area5~\cite{armeni20163d} and ScanNet v2 validation set~\cite{dai2017scannet} (\S\!~\ref{sec:downstream2}).} We report mIoU (\%) here.}
\label{tab:segmentation1}
\centering
\small
\setlength\tabcolsep{4pt}
\renewcommand\arraystretch{1.1}
\resizebox{\linewidth}{!}{
\begin{tabular}{rccc}
\toprule[0.95pt]
Methods & Backbone & S3DIS Area5 &ScanNet v2\\  
\midrule[0.6pt]
\multicolumn{4}{c}{\textit{Supervised Learning Only}}\\
\midrule[0.6pt]
PointNet++~\cite{qi2017pointnet++} & PointNet++ & 55.3 & 57.9\\
MinkNet~\cite{choy20194d} & SR-UNet & 68.2 & 70.3\\
\rowcolor{mygray}
MH-V (Ours) & MH-V & 70.3 & 71.4\\
\midrule[0.6pt]
\multicolumn{4}{c}{\textit{Pre-training on ScanNet}}\\
\midrule[0.6pt]
PointContrast~\cite{xie2020pointcontrast}& SR-UNet &70.9 &74.1 \\ 
DepthContrast~\cite{zhang2021self} & SR-UNet & 71.5 & 71.2 \\
OcCo~\cite{wang2021unsupervised} &DGCNN &58.0 &-\\
PointClustering~\cite{long2023pointclustering} &PointNet++ &61.2 & 62.6\\
PointClustering~\cite{long2023pointclustering} &Transformer & 65.6 & 65.8\\
PointClustering~\cite{long2023pointclustering} & SR-UNet & 73.2 & 75.5\\
\midrule[0.6pt]
\multicolumn{4}{c}{\textit{Pre-training on ShapeNet / UHD and LHD}}\\
\midrule[0.6pt]
\rowcolor{mygray}
MH-VS (Ours) & MH-V & 72.7 & 74.4\\
\rowcolor{mygray}
MH-VH (Ours) & MH-V & \bf{74.1} & \bf{75.8}\\
\bottomrule[0.95pt]
\end{tabular}}
\end{minipage}
\hfill
\begin{minipage}[t]{0.49\linewidth}
\captionsetup{font=small}
\caption{\small
\textbf{Segmentation results on Semantic- 
KITTI validation set~\cite{behley2019semantickitti} and Synthia4D test set~\cite{ros2016synthia} (\S\!~\ref{sec:downstream2}).} We report mIoU (\%) here.}
\label{tab:segmentation2}
\centering
\small
\setlength\tabcolsep{4pt}
\renewcommand\arraystretch{1.1}
\resizebox{\linewidth}{!}{
\begin{tabular}{rccc}
\toprule[0.95pt]
Methods & Backbone & SemanticKITTI &Synthia4D\\  
\midrule[0.6pt]
\multicolumn{4}{c}{\textit{Supervised Learning Only}}\\
\midrule[0.6pt]
PointNet++~\cite{qi2017pointnet++} & PointNet++ & - & 79.4\\
MinkNet~\cite{choy20194d,tang2020searching} & SR-UNet & 61.1 & 79.8\\
SPVNAS~\cite{tang2020searching} & SPVCNN & 64.7 & - \\
\rowcolor{mygray}
MH-V (Ours) & MH-V & 67.1 & 81.4\\
\midrule[0.6pt]
\multicolumn{4}{c}{\textit{Pre-training on ScanNet}}\\
\midrule[0.6pt]
PointContrast~\cite{xie2020pointcontrast}& SR-UNet & - &83.1 \\ 
DepthContrast~\cite{zhang2021self} & SR-UNet & - & 81.3 \\
\midrule[0.6pt]
\multicolumn{4}{c}{\textit{Pre-training on ShapeNet}}\\
\midrule[0.6pt]
PointDif~\cite{zheng2023point}& SR-UNet & 71.3 & -\\
\rowcolor{mygray}
MH-VS (Ours) & MH-V & 70.8 & 82.4\\
\midrule[0.6pt]
\multicolumn{4}{c}{\textit{Pre-training on UHD and LHD}}\\
\midrule[0.6pt]
\rowcolor{mygray}
MH-VH (Ours) & MH-V & \bf{71.5} & \bf{84.2}\\
\bottomrule[0.95pt]
\end{tabular}}
\end{minipage}
\end{table}

\subsection{Scene Level Downstream Tasks}
\label{sec:downstream2}
We perform fine-tuning for MH-VS and  MH-VH on five scene-level downstream tasks. Each scene-level downstream task has hundreds of thousands of input points.

\noindent\textbf{Semantic Segmentation on Indoor Scene.}~We proceed with the evaluation of MH-V in the indoor semantic segmentation task, involving the classification of points within 3D scenes into distinct categories. This research involves pre-training on pseudo scenes and finetuning on the Stanford Large-Scale 3D Indoor Spaces (S3DIS) dataset~\cite{armeni20163d} and ScanNet v2 validation set~\cite{dai2017scannet}, respectively. They are based on the standard settings~\cite{choy20194d,xie2020pointcontrast,long2023pointclustering}. S3DIS encompasses 3D scans of six expansive indoor areas. Following the prior method~\cite{choy20194d}, Area 5 is designated as the test set. There are approximately 204 samples in the training set. Each training sample contains an average of $\sim$956K points. ScanNet v2 comprises richly annotated 3D indoor scenes, encompassing 1.5K scenes from hundreds of distinct rooms. There are 1,201 scenes for training. The mean Intersection over Union (mIoU) of various approaches is summarized in Tab.\!~\ref{tab:segmentation1}. Even with S2SS, transferring features from the pseudo scene to S3DIS and ScanNet v2 presents a considerable challenge. While UHD includes specific objects outlined in S3DIS, discrepancies arise in terms of occlusion, object scale, and scene magnitude. However, when compared to pre-training methods on ScanNet, such as OcCo~\cite{wang2021unsupervised} (with DGCNN as the backbone), and PointClustering~\cite{long2023pointclustering} (with PointNet++, PointViT, and SR-UNet as backbones), MH-V shows significant improvement. It also outperforms pre-training methods~\cite{xie2020pointcontrast,zhang2021self} that use SR-UNet as the backbone. These results conclusively affirm that MH-V derives substantial advantages from the acquisition of high-resolution features across various scales. The high-resolution representation effectively harnesses the semantics within point cloud data, thereby bolstering the network's proficiency in semantic segmentation.

\noindent\textbf{Semantic Segmentation on Outdoor Scene.}~We also conduct experiments on SemanticKITTI~\cite{behley2019semantickitti} and Synthia4D dataset~\cite{ros2016synthia} for the outdoor semantic segmentation task. SemanticKITTI is a large-scale driving-scene dataset, containing 43K scans with point-wise annotation. We use sequences 00 to 10 for training, and 08 is left for validation. Synthia4D is a large synthetic dataset depicting driving scenarios. We adhere to the train/validation/test split as defined by \cite{choy20194d}. Our implementation does not involve temporal learning. mIoU of different methods is summarized in Tab.\!~\ref{tab:segmentation2}. Fine-tuning on them poses similar challenges as previously mentioned. However, we significantly mitigate the differences between shape data and scene data by employing synthesized pseudo scenes. In comparison to the methods pre-trained on the ScanNet and ShapeNet datasets, our approach has also achieved encouraging results. This indicates that the representations learned in pseudo scenes still enhance the generalization for segmentation in both real-world and synthetic outdoor scenes.

\begin{table}[t]
\captionsetup{font=small}
\caption{\small
\textbf{3D object detection on ScanNet v2~\cite{dai2017scannet} (\S\!~\ref{sec:downstream2}).} We report mAP (\%). SR-UNet stands for Sparse Residual U-Net~\cite{xie2020pointcontrast}.}
\label{tab:detection}
\begin{minipage}[t]{.49\textwidth}
\centering
\small
\setlength\tabcolsep{4.8pt}
\renewcommand\arraystretch{1.1}
\resizebox{\linewidth}{!}{
\begin{tabular}{rccc}
\toprule[0.95pt]
Methods &Backbone & mAP@0.25 & mAP@0.5\\
\midrule[0.6pt]
\multicolumn{4}{c}{\textit{Supervised Learning Only}}\\
\midrule[0.6pt]
VoteNet~\cite{qi2019deep}& PointNet++ &58.6 &33.5 \\
3DETR~\cite{misra2021end}& Transformer &62.1 &37.9 \\
\rowcolor{mygray} MH-V (Ours) & MH-V & 62.9  &39.8 \\
\midrule[0.6pt]
\multicolumn{4}{c}{\textit{Pre-training on ShapeNet}}\\
\midrule[0.6pt]
TAP~\cite{wang2023take} & Transformer & 63.0 & 41.4 \\
PointDif~\cite{zheng2023point} & Transformer & - & 43.7\\
\rowcolor{mygray} MH-VS (Ours) & MH-V & 63.4  &42.1\\
\bottomrule[0.95pt]
\end{tabular}}
\end{minipage}
\hfill
\begin{minipage}[t]{0.49\linewidth}
\centering
\small
\setlength\tabcolsep{4.8pt}
\renewcommand\arraystretch{1.1}
\resizebox{\linewidth}{!}{
\begin{tabular}{rccc}
\toprule[0.95pt]
Methods &Backbone & mAP@0.25 & mAP@0.5\\
\midrule[0.6pt]
\multicolumn{4}{c}{\textit{Pre-training on ScanNet}}\\
\midrule[0.6pt]
PointContrast~\cite{xie2020pointcontrast}& SR-UNet &59.2 &38.0 \\
STRL~\cite{huang2021spatio}& PointNet++ &59.5 &38.4 \\
Point-BERT~\cite{yu2022point} & Transformer & 61.0 &38.3 \\
DepthContrast~\cite{zhang2021self} & PointNet++ &64.0 & 42.9\\
MaskPoint~\cite{liu2022masked}& Transformer & 64.2 & 42.1 \\
\midrule[0.6pt]
\multicolumn{4}{c}{\textit{Pre-training on UHD and LHD}}\\
\midrule[0.6pt]
\rowcolor{mygray} MH-VH (Ours) & MH-V &\textbf{64.8}  &\textbf{43.9} \\
\bottomrule[0.95pt]
\end{tabular}}
\end{minipage}
\end{table}

\noindent\textbf{3D Object Detection on ScanNet v2.}~To further evaluate MH-V on 3D object detection, we utilize MH-V as the backbone on the ScanNet v2 dataset~\cite{dai2017scannet}. Our methodology adheres to the training/validation split and 18 classes outlined in VoteNet~\cite{qi2019deep}. We also follow VoteNet~\cite{qi2019deep} and switch the original backbone network with MH-V without other modifications to the detection module. As reported in Tab.\!~\ref{tab:detection}, MH-VH outperforms MH-V by 4.1\% mAP@0.5. Previous methods~\cite{xie2020pointcontrast, huang2021spatio, yu2022point, zhang2021self, liu2022masked} conduct pre-training on large-scale scene datasets like ScanNet. In contrast, we conduct pre-training on pseudo scenes originally from individual objects. The experiments also show the effectiveness of MH-V over SR-UNet, as MH-V learns multi-scale point cloud encoding by mapping scale features to high-resolution features. This demonstrates that the representations learned from multiple objects can be successfully transferred to large-scale datasets and enhance the performance of scene-level tasks.

\subsection{Ablation Study}
\label{sec:ablation}

\noindent\textbf{Comparsion to Training-from-scratch Baseline.}~We first compare the training-from-scratch baseline and the fine-tuned network on ScanObjectNN (PB\_T50\_RS), ShapeNetPart, and S3DIS Area5, respectively. Specifically, MH-PS and MH-VS are pre-trained on ShapeNet and fine-tuned on the target dataset. The upper part of Tab.\!~\ref{tab:abl1} details the model configurations. \textbf{\texttt{Model\!\! A}}, \textbf{\texttt{B}}, \textbf{\texttt{D}}, and \textbf{\texttt{E}} are employed as training-from-scratch baselines. The backbone of \textbf{\texttt{Model\!\! A}} is PointNet++ with an encoder on ScanObjectNN, while it uses PointNet++ with both encoder and decoder on ShapeNetPart. \textbf{\texttt{Model\!\! C}}, \textbf{\texttt{F}} leverages pseudo scenes synthesized through S2SS for pre-training data, with PPC as the loss function. In the lower part of Tab.\!~\ref{tab:abl1}, we present the performance on shape-level tasks (ScanObjectNN, ShapeNetPart) and scene-level tasks (S3DIS Area5). On ScanObjectNN, \textbf{\texttt{Model\!\! B}} aggregates deep semantic features at multiple scales, outperforming \textbf{\texttt{Model\!\! A}}. This proves that MH-P is a strong backbone. For high-resolution tasks, we benefit from the design of multi-scale feature aggregation with high-resolution mapping. Both \textbf{\texttt{Model\!\! B}} and \textbf{\texttt{E}} exhibit better performance than \textbf{\texttt{Model\!\! A}} and \textbf{\texttt{D}}. Compared to training-from-scratch baselines, \textbf{\texttt{Model\!\! C}} achieves improvements of 1.9\% and 1.1\% on ScanObjectNN and ShapeNetPart, respectively. \textbf{\texttt{Model\!\! F}} demonstrated a 2.4\% improvement on S3DIS compared to \textbf{\texttt{Model\!\! E}}. This highlights the efficacy of our proposed pre-training strategy, \ie, S2SS and PPC, for high-resolution tasks.

\begin{table}[t]
\begin{minipage}[t]{.49\textwidth}
\captionsetup{font=small}
\caption{\small
\textbf{Ablations for different designs (\S\!~\ref{sec:ablation}).} We report Overall Accuracy (\%) on ScanObjectNN (PB\_T50\_RS), Instance mIoU (\%) on ShapeNetPart, and mIoU (\%) on S3DIS Area5.}
\label{tab:abl1}
\begin{subtable}[t]{\linewidth}
\centering
\small
\setlength\tabcolsep{9.5pt}
\renewcommand\arraystretch{1.1}
\resizebox{\linewidth}{!}{
\begin{tabular}{rccc}
\toprule[0.95pt]
Models & Pre-training & Backbone & Data Type \\
\midrule[0.6pt]
\textbf{\texttt{Model A}}  & {\color{gray}Scratch} & PointNet++ & Shape\\
\textbf{\texttt{Model B}}  & {\color{gray}Scratch} & MH-P & Shape\\
\textbf{\texttt{Model C}}  & \checkmark&  MH-P & Shape\\
\cdashline{1-4}[1pt/1pt]
\textbf{\texttt{Model D}}  & {\color{gray}Scratch}& SR-UNet & Scene\\
\textbf{\texttt{Model E}} & {\color{gray}Scratch}&  MH-V & Scene \\
\textbf{\texttt{Model F}}  & \checkmark&  MH-V & Scene\\
\bottomrule[0.95pt]
\end{tabular}}
\end{subtable}
  \hfill
\begin{subtable}[t]{\linewidth}
\centering
\small
\setlength\tabcolsep{5.6pt}
\renewcommand\arraystretch{1.1}
\resizebox{\linewidth}{!}{
\begin{tabular}{rccc}
\toprule[0.95pt]
Models  & ScanObjectNN & ShapeNetPart & S3DIS Area5\\
\midrule[0.6pt]
\textbf{\texttt{Model A}}  & 77.9 & 85.1 & - \\
\textbf{\texttt{Model B}}  & 85.9  & 86.0 &  - \\
\rowcolor{mygray}
\textbf{\texttt{Model C}}  &  87.8 & 87.1 &  - \\
\cdashline{1-4}[1pt/1pt]
\textbf{\texttt{Model D}}  & - & - & 68.2 \\
\textbf{\texttt{Model E}}  &  - & - &  70.3 \\
\rowcolor{mygray}
\textbf{\texttt{Model F}}  &  - & - &  72.7 \\
\bottomrule[0.95pt]
\end{tabular}}
\end{subtable}
\end{minipage}
\hfill
\begin{minipage}[t]{0.49\linewidth}
\captionsetup{font=small}
\caption{\small
\textbf{Ablations for network design (\S\!~\ref{sec:ablation}).} We report Overall Accuracy (\%) on ScanObjectNN (PB\_T50\_RS), Instance mIoU (\%) on ShapeNetPart, and mIoU (\%) on S3DIS Area5.}
\label{tab:abl2}
\begin{subtable}[t]{\linewidth}
\centering
\small
\setlength\tabcolsep{7.5pt}
\renewcommand\arraystretch{0.9}
\resizebox{\linewidth}{!}{
\begin{tabular}{rccccc}
\toprule[0.95pt]
Models & Backbone & $S_1$ & $S_2$ & $S_3$ & $S_4$ \\
\midrule[0.6pt]
\textbf{\texttt{Model G}}& MH-P & \checkmark/\checkmark & \checkmark/\checkmark &  & \\
\textbf{\texttt{Model H}}& MH-P  & \checkmark/\checkmark & \checkmark/\checkmark & \checkmark/\checkmark & \\
\textbf{\texttt{Model I}}& MH-P  & \checkmark/\checkmark & \checkmark/\checkmark & \checkmark/\checkmark & \checkmark/\checkmark \\
\textbf{\texttt{Model J}}& MH-P  & \checkmark/\hfill\hfill\hfill\hfill &  \checkmark/\hfill\hfill\hfill\hfill & \checkmark/\checkmark & \checkmark/\checkmark \\
\textbf{\texttt{Model K}}& MH-P  & \checkmark/\hfill\hfill\hfill\hfill & \checkmark/\hfill\hfill\hfill\hfill & \checkmark/\hfill\hfill\hfill\hfill &  \checkmark/\checkmark \\
\cdashline{1-6}[1pt/1pt]
\textbf{\texttt{Model L}} & MH-V & \checkmark/\checkmark & \checkmark/\checkmark &  & \\
\textbf{\texttt{Model M}} & MH-V & \checkmark/\checkmark & \checkmark/\checkmark & \checkmark/\checkmark & \\
\textbf{\texttt{Model N}} & MH-V & \checkmark/\checkmark & \checkmark/\checkmark & \checkmark/\checkmark & \checkmark/\checkmark \\
\textbf{\texttt{Model O}}  & MH-V & \checkmark/\hfill\hfill\hfill\hfill &  \checkmark/\hfill\hfill\hfill\hfill & \checkmark/\checkmark & \checkmark/\checkmark \\
\textbf{\texttt{Model P}} & MH-V & \checkmark/\hfill\hfill\hfill\hfill & \checkmark/\hfill\hfill\hfill\hfill & \checkmark/\hfill\hfill\hfill\hfill &  \checkmark/\checkmark \\
\bottomrule[0.95pt]
\end{tabular}}
\end{subtable}
  \hfill
\begin{subtable}[t]{\linewidth}
\centering
\small
\setlength\tabcolsep{5.6pt}
\renewcommand\arraystretch{0.9}
\resizebox{\linewidth}{!}{
\begin{tabular}{rccc}
\toprule[0.95pt]
Models  & ScanObjectNN & ShapeNetPart & S3DIS Area5\\
\midrule[0.6pt]
\textbf{\texttt{Model G}}  & 82.3 & 85.4 & - \\
\textbf{\texttt{Model H}}  & 85.6  & 86.3 &  - \\
\rowcolor{mygray}
\textbf{\texttt{Model I}}  & 87.8 & 87.1& - \\
\textbf{\texttt{Model J}}  & 87.4 & 86.8  &  - \\
\textbf{\texttt{Model K}}  &  87.1 & 86.6 &  - \\
\cdashline{1-4}[1pt/1pt]
\textbf{\texttt{Model L}}  & - & - & 62.3 \\
\textbf{\texttt{Model M}}  &  - & - &  68.4 \\
\rowcolor{mygray}
\textbf{\texttt{Model N}}  &  - & - &  72.7 \\
\textbf{\texttt{Model O}}  &  - & - &  72.0 \\
\textbf{\texttt{Model P}}  &  - & - &  71.6 \\
\bottomrule[0.95pt]
\end{tabular}}
\end{subtable}
\end{minipage}
\end{table}

\noindent\textbf{Network Design.}~Then we present the performance of models that have different network designs. We have investigated four scales, denoted as $S_1\!=\!2$, $S_2\!=\!4$, $S_3\!=\!8$, and $S_4\!=\!16$. The upper part of Tab.\!~\ref{tab:abl2} details the configurations of scales. Specifically, whether to use a certain scale and whether to integrate its feature into the final output. \textbf{\texttt{Model\!\! G}}-\textbf{\texttt{K}} are pre-trained on ShapeNet\!~\cite{chang2015shapenet} with PPC. \textbf{\texttt{Model\!\! L}}-\textbf{\texttt{P}} synthesize pseudo scene with S2SS on ShapeNet\!~\cite{chang2015shapenet} under the supervision of PPC. In the lower part of Tab.\!~\ref{tab:abl2}, we report the performance. MH-P and MH-V achieve optimal performance when using four scales. Comparing \textbf{\texttt{Model\!\! G}}-\textbf{\texttt{K}} and \textbf{\texttt{Model\!\! L}}-\textbf{\texttt{P}}, we find that integrating high-resolution features from multiple scales into the final output results in improved performance. This aligns with the key advantages discussed in \S\!~\ref{sec:introduction}.

\noindent\textbf{Pre-Training Data for Scene-Level Task.} Furthermore, we have conducted research on pre-training data for the scene-level task (S3DIS Area5). \textbf{\texttt{Model\!\! Q}}, \textbf{\texttt{R}}, \textbf{\texttt{S}}, and \textbf{\texttt{T}} have all used MH-V as the backbone and have been pre-trained on ShapeNet (shape data), pseudo scenes made of ShapeNet (scene data), ScanNet from PointContrast\!~\cite{xie2020pointcontrast} (scene data), and pseudo scenes made of UHD (scene data), respectively. \textbf{\texttt{Model\!\! Q}} shows only marginal improvement compared to \textbf{\texttt{Model\!\! E}} (refer to Tab.\!~{\color{red}6}). Pre-training on shape-level data exhibit minimal potential in downstream scene-level tasks. However, \textbf{\texttt{Model\!\! R}} demonstrates significantly enhanced performance compared to \textbf{\texttt{Model\!\! Q}}, indicating the effectiveness of S2SS. Additionally, \textbf{\texttt{Model\!\! T}} even outperforms \textbf{\texttt{Model\!\! S}} (pre-training on ScanNet) with larger shape-level datasets (\ie, UHD and LHD\!~\cite{chen2023pointgpt}). Considering the time-consuming preprocessing required for generating matched pairs in PointContrast\!~\cite{xie2020pointcontrast}, our approach can be integrated into the training process to obtain matched pairs in running time, which is highly valuable.

\begin{table}[t]
\begin{minipage}[t]{0.49\linewidth}
\captionsetup{font=small}
\caption{\small
\textbf{Ablation study on pre-training data for the scene-level task (\S\!~\ref{sec:ablation}).} We report mIoU (\%) on S3DIS Area5.}
\label{tab:abl3}
\centering
\small
\setlength\tabcolsep{2.6pt}
\renewcommand\arraystretch{1.1}
\resizebox{\linewidth}{!}{
\begin{tabular}{rcccc}
\toprule[0.95pt]
Models & Backbone & Data Type & Source & S3DIS Area5 \\
\midrule[0.6pt]
\textbf{\texttt{Model Q}}& MH-V  & Shape & ShapeNet\!~\cite{chang2015shapenet} & 70.5 \\
\textbf{\texttt{Model R}}& MH-V  & Scene & ShapeNet\!~\cite{chang2015shapenet} & 72.7\\
\textbf{\texttt{Model S}}& MH-V  & Scene & ScanNet\!~\cite{xie2020pointcontrast} & 73.5\\
\rowcolor{mygray}
\textbf{\texttt{Model T}}& MH-V  & Scene & UHD\!~\cite{chen2023pointgpt} & 74.1\\
\bottomrule[0.95pt]
\end{tabular}}
\end{minipage}
\hfill
\begin{minipage}[t]{.49\textwidth}
\captionsetup{font=small}
\caption{\small
\textbf{Ablation study for $M$ (\S\!~\ref{sec:ablation}).} We report OA (\%) on ScanObjectNN, Inst. mIoU (\%) on ShapeNetPart, and mIoU (\%) on S3DIS.}
\label{tab:abl4}
\centering
\small
\setlength\tabcolsep{8.4pt}
\renewcommand\arraystretch{1.0}
\resizebox{\linewidth}{!}{
\begin{tabular}{rccc}
\toprule[0.95pt]
$M$ & ScanObjectNN & ShapeNetPart & S3DIS Area5 \\
\midrule[0.6pt]
1  & 86.8 & 86.2 & 70.5 \\
2  & 87.4 & 86.7 & 71.1\\
4  &\cellcolor{mygray}87.8 &\cellcolor{mygray}87.1 & 71.9\\
6  & 87.7 & 86.9 & \cellcolor{mygray}72.7\\
8  & 87.7 & 86.7 & 72.5\\
10 & 87.6 & 87.0 & 72.6\\
\bottomrule[0.95pt]
\end{tabular}}
\end{minipage}
\end{table}

\noindent\textbf{Number of Shapes $M$.}~Finally, we conduct ablation experiments to explore $M$ in S2SS on three downstream tasks. The pre-training settings, including pre-training data, S2SS, MH-P/V, loss function, \etc., remain consistent with \textbf{\texttt{Model\!\! I}} and \textbf{\texttt{N}}. In downstream shape-level tasks, the pre-trained model demonstrates its peak performance at $M\!=\!4$. In contrast, for scene-level segmentation, the optimal performance of the pre-trained model is observed at $M\!=\!6$. When it comes to larger values of $M$, neither \textbf{\texttt{Model\!\! I}} nor \textbf{\texttt{N}} show improved performance. Intuitively, we attribute this to the number of matched pairs utilized for the contrastive loss. However, indiscriminately increasing the number of matched pairs entails in increased resource and time consumption~\cite{xie2020pointcontrast}, hence we empirically set $M\!=\!4$ for \textbf{\texttt{Model\!\! I}} and $M\!=\!6$ for \textbf{\texttt{Model\!\! N}}.

\section{Conclusion}
In summary, our research introduces S2S, a novel 3D scene representation learning method. By leveraging readily available 3D shape datasets and introducing innovative architectural designs, data strategies, and loss functions, S2S overcomes the \textit{data desert} issue and bridges the gap between shape-level and scene-level datasets. This approach demonstrates remarkable adaptability and transferability across both shape-level and scene-level datasets, as evidenced by the impressive performance achieved on eight well-established benchmarks. S2S represents a significant step forward in addressing the challenges faced by current 3D SSL for scene understanding. However, despite its notable achievements, challenges still lie ahead, particularly in extending the application of S2S to broader 3D contexts, such as open-world scene understanding. Expanding S2S's capabilities to encompass these more complex scenarios will require further research and innovation. We intend to delve into these challenges in our future research.

%
%
\bibliographystyle{splncs04_unsort}
\bibliography{egbib}

\begin{thebibliography}{10}
\providecommand{\url}[1]{\texttt{#1}}
\providecommand{\urlprefix}{URL }
\providecommand{\doi}[1]{https://doi.org/#1}

\bibitem{radford2018improving}
Radford, A., Narasimhan, K., Salimans, T., Sutskever, I., et~al.: Improving
  language understanding by generative pre-training. In: preprint. OpenAI
  (2018)

\bibitem{devlin2018bert}
Devlin, J., Chang, M.W., Lee, K., Toutanova, K.: Bert: Pre-training of deep
  bidirectional transformers for language understanding. arXiv preprint
  arXiv:1810.04805  (2018)

\bibitem{brown2020language}
Brown, T., Mann, B., Ryder, N., Subbiah, M., Kaplan, J.D., Dhariwal, P.,
  Neelakantan, A., Shyam, P., Sastry, G., Askell, A., et~al.: Language models
  are few-shot learners. In: NeurIPS (2020)

\bibitem{wei2022chain}
Wei, J., Wang, X., Schuurmans, D., Bosma, M., Xia, F., Chi, E., Le, Q.V., Zhou,
  D., et~al.: Chain-of-thought prompting elicits reasoning in large language
  models. In: NeurIPS (2022)

\bibitem{ouyang2022training}
Ouyang, L., Wu, J., Jiang, X., Almeida, D., Wainwright, C., Mishkin, P., Zhang,
  C., Agarwal, S., Slama, K., Ray, A., et~al.: Training language models to
  follow instructions with human feedback. In: NeurIPS (2022)

\bibitem{he2020momentum}
He, K., Fan, H., Wu, Y., Xie, S., Girshick, R.: Momentum contrast for
  unsupervised visual representation learning. In: CVPR (2020)

\bibitem{he2022masked}
He, K., Chen, X., Xie, S., Li, Y., Doll{\'a}r, P., Girshick, R.: Masked
  autoencoders are scalable vision learners. In: CVPR (2022)

\bibitem{radford2021learning}
Radford, A., Kim, J.W., Hallacy, C., Ramesh, A., Goh, G., Agarwal, S., Sastry,
  G., Askell, A., Mishkin, P., Clark, J., et~al.: Learning transferable visual
  models from natural language supervision. In: ICML (2021)

\bibitem{rombach2022high}
Rombach, R., Blattmann, A., Lorenz, D., Esser, P., Ommer, B.: High-resolution
  image synthesis with latent diffusion models. In: CVPR (2022)

\bibitem{alayrac2022flamingo}
Alayrac, J.B., Donahue, J., Luc, P., Miech, A., Barr, I., Hasson, Y., Lenc, K.,
  Mensch, A., Millican, K., Reynolds, M., et~al.: Flamingo: a visual language
  model for few-shot learning. In: NeurIPS (2022)

\bibitem{dong2022autoencoders}
Dong, R., Qi, Z., Zhang, L., Zhang, J., Sun, J., Ge, Z., Yi, L., Ma, K.:
  Autoencoders as cross-modal teachers: Can pretrained 2d image transformers
  help 3d representation learning? arXiv preprint arXiv:2212.08320  (2022)

\bibitem{geiger2013vision}
Geiger, A., Lenz, P., Stiller, C., Urtasun, R.: Vision meets robotics: The
  kitti dataset  (2013)

\bibitem{dai2017scannet}
Dai, A., Chang, A.X., Savva, M., Halber, M., Funkhouser, T., Nie{\ss}ner, M.:
  Scannet: Richly-annotated 3d reconstructions of indoor scenes. In: CVPR
  (2017)

\bibitem{armeni20163d}
Armeni, I., Sener, O., Zamir, A.R., Jiang, H., Brilakis, I., Fischer, M.,
  Savarese, S.: 3d semantic parsing of large-scale indoor spaces. In: CVPR
  (2016)

\bibitem{wu20153d}
Wu, Z., Song, S., Khosla, A., Yu, F., Zhang, L., Tang, X., Xiao, J.: 3d
  shapenets: A deep representation for volumetric shapes. In: CVPR (2015)

\bibitem{uy2019revisiting}
Uy, M.A., Pham, Q.H., Hua, B.S., Nguyen, T., Yeung, S.K.: Revisiting point
  cloud classification: A new benchmark dataset and classification model on
  real-world data. In: ICCV (2019)

\bibitem{github}
{GitHub}. \url {https://github.com/}

\bibitem{Thingiverse}
{Thingiverse}. \url {https://www.thingiverse.com/}

\bibitem{Sketchfab}
{Sketchfab}. \url {https://sketchfab.com/}

\bibitem{Polycam}
{Polycam}. \url {https://poly.cam/}

\bibitem{Smithsonian3DDigitization}
{Smithsonian 3D Digitization}. \url {https://3d.si.edu//}

\bibitem{threestudio2023}
Guo, Y.C., Liu, Y.T., Shao, R., Laforte, C., Voleti, V., Luo, G., Chen, C.H.,
  Zou, Z.X., Wang, C., Cao, Y.P., Zhang, S.H.: threestudio: A unified framework
  for 3d content generation.
  \url{https://github.com/threestudio-project/threestudio} (2023)

\bibitem{wu2023multiview}
Wu, C.Y., Johnson, J., Malik, J., Feichtenhofer, C., Gkioxari, G.: Multiview
  compressive coding for 3{D} reconstruction. arXiv:2301.08247  (2023)

\bibitem{yu2022point}
Yu, X., Tang, L., Rao, Y., Huang, T., Zhou, J., Lu, J.: Point-bert:
  Pre-training 3d point cloud transformers with masked point modeling. In: CVPR
  (2022)

\bibitem{liu2022masked}
Liu, H., Cai, M., Lee, Y.J.: Masked discrimination for self-supervised learning
  on point clouds. In: ECCV (2022)

\bibitem{pang2022masked}
Pang, Y., Wang, W., Tay, F.E., Liu, W., Tian, Y., Yuan, L.: Masked autoencoders
  for point cloud self-supervised learning. In: ECCV (2022)

\bibitem{zhang2022point}
Zhang, R., Guo, Z., Gao, P., Fang, R., Zhao, B., Wang, D., Qiao, Y., Li, H.:
  Point-m2ae: multi-scale masked autoencoders for hierarchical point cloud
  pre-training. In: NeurIPS (2022)

\bibitem{chen2023pointgpt}
Chen, G., Wang, M., Yang, Y., Yu, K., Yuan, L., Yue, Y.: Pointgpt:
  Auto-regressively generative pre-training from point clouds. In: NeurIPS
  (2023)

\bibitem{xie2020pointcontrast}
Xie, S., Gu, J., Guo, D., Qi, C.R., Guibas, L., Litany, O.: Pointcontrast:
  Unsupervised pre-training for 3d point cloud understanding. In: ECCV (2020)

\bibitem{zhang2021self}
Zhang, Z., Girdhar, R., Joulin, A., Misra, I.: Self-supervised pretraining of
  3d features on any point-cloud. In: ICCV (2021)

\bibitem{meng2021towards}
Meng, Q., Wang, W., Zhou, T., Shen, J., Jia, Y., Van~Gool, L.: Towards a weakly
  supervised framework for 3d point cloud object detection and annotation. IEEE
  TPAMI  \textbf{44}(8),  4454--4468 (2021)

\bibitem{meng2020weakly}
Meng, Q., Wang, W., Zhou, T., Shen, J., Van~Gool, L., Dai, D.: Weakly
  supervised 3d object detection from lidar point cloud. In: ECCV (2020)

\bibitem{yin2022semi}
Yin, J., Fang, J., Zhou, D., Zhang, L., Xu, C.Z., Shen, J., Wang, W.:
  Semi-supervised 3d object detection with proficient teachers. In: ECCV (2022)

\bibitem{yin2024fusion}
Yin, J., Shen, J., Chen, R., Li, W., Yang, R., Frossard, P., Wang, W.:
  Is-fusion: Instance-scene collaborative fusion for multimodal 3d object
  detection. In: CVPR (2024)

\bibitem{choy2019fully}
Choy, C., Park, J., Koltun, V.: Fully convolutional geometric features. In:
  ICCV (2019)

\bibitem{yi2016scalable}
Yi, L., Kim, V.G., Ceylan, D., Shen, I.C., Yan, M., Su, H., Lu, C., Huang, Q.,
  Sheffer, A., Guibas, L.: A scalable active framework for region annotation in
  3d shape collections. ACM TOG  \textbf{35}(6),  1--12 (2016)

\bibitem{behley2019semantickitti}
Behley, J., Garbade, M., Milioto, A., Quenzel, J., Behnke, S., Stachniss, C.,
  Gall, J.: Semantickitti: A dataset for semantic scene understanding of lidar
  sequences. In: ICCV (2019)

\bibitem{ros2016synthia}
Ros, G., Sellart, L., Materzynska, J., Vazquez, D., Lopez, A.M.: The synthia
  dataset: A large collection of synthetic images for semantic segmentation of
  urban scenes. In: CVPR (2016)

\bibitem{qi2017pointnet}
Qi, C.R., Su, H., Mo, K., Guibas, L.J.: Pointnet: Deep learning on point sets
  for 3d classification and segmentation. In: CVPR (2017)

\bibitem{qi2017pointnet++}
Qi, C.R., Yi, L., Su, H., Guibas, L.J.: Pointnet++: Deep hierarchical feature
  learning on point sets in a metric space. In: NeurIPS (2017)

\bibitem{li2018pointcnn}
Li, Y., Bu, R., Sun, M., Wu, W., Di, X., Chen, B.: Pointcnn: Convolution on
  x-transformed points. In: NeurIPS (2018)

\bibitem{yang2019std}
Yang, Z., Sun, Y., Liu, S., Shen, X., Jia, J.: Std: Sparse-to-dense 3d object
  detector for point cloud. In: ICCV (2019)

\bibitem{ma2021rethinking}
Ma, X., Qin, C., You, H., Ran, H., Fu, Y.: Rethinking network design and local
  geometry in point cloud: A simple residual mlp framework. In: ICLR (2021)

\bibitem{qian2022pointnext}
Qian, G., Li, Y., Peng, H., Mai, J., Hammoud, H., Elhoseiny, M., Ghanem, B.:
  Pointnext: Revisiting pointnet++ with improved training and scaling
  strategies. In: NeurIPS (2022)

\bibitem{feng2024interpretable3d}
Feng, T., Quan, R., Wang, X., Wang, W., Yang, Y.: Interpretable3d: An ad-hoc
  interpretable classifier for 3d point clouds. In: AAAI (2024)

\bibitem{wu2018squeezeseg}
Wu, B., Wan, A., Yue, X., Keutzer, K.: Squeezeseg: Convolutional neural nets
  with recurrent crf for real-time road-object segmentation from 3d lidar point
  cloud. In: ICRA (2018)

\bibitem{zhang2020polarnet}
Zhang, Y., Zhou, Z., David, P., Yue, X., Xi, Z., Gong, B., Foroosh, H.:
  Polarnet: An improved grid representation for online lidar point clouds
  semantic segmentation. In: CVPR (2020)

\bibitem{xu2018spidercnn}
Xu, Y., Fan, T., Xu, M., Zeng, L., Qiao, Y.: Spidercnn: Deep learning on point
  sets with parameterized convolutional filters. In: ECCV (2018)

\bibitem{tatarchenko2018tangent}
Tatarchenko, M., Park, J., Koltun, V., Zhou, Q.Y.: Tangent convolutions for
  dense prediction in 3d. In: CVPR (2018)

\bibitem{choy20194d}
Choy, C., Gwak, J., Savarese, S.: 4d spatio-temporal convnets: Minkowski
  convolutional neural networks. In: CVPR (2019)

\bibitem{yan2018second}
Yan, Y., Mao, Y., Li, B.: Second: Sparsely embedded convolutional detection.
  Sensors  \textbf{18}(10), ~3337 (2018)

\bibitem{riegler2017octnet}
Riegler, G., Osman~Ulusoy, A., Geiger, A.: Octnet: Learning deep 3d
  representations at high resolutions. In: CVPR (2017)

\bibitem{graham20183d}
Graham, B., Engelcke, M., van~der Maaten, L.: 3d semantic segmentation with
  submanifold sparse convolutional networks. In: CVPR (2018)

\bibitem{klokov2017escape}
Klokov, R., Lempitsky, V.: Escape from cells: Deep kd-networks for the
  recognition of 3d point cloud models. In: ICCV (2017)

\bibitem{zhu2021cylindrical}
Zhu, X., Zhou, H., Wang, T., Hong, F., Ma, Y., Li, W., Li, H., Lin, D.:
  Cylindrical and asymmetrical 3d convolution networks for lidar segmentation.
  In: CVPR (2021)

\bibitem{afham2022crosspoint}
Afham, M., Dissanayake, I., Dissanayake, D., Dharmasiri, A., Thilakarathna, K.,
  Rodrigo, R.: Crosspoint: Self-supervised cross-modal contrastive learning for
  3d point cloud understanding. In: CVPR (2022)

\bibitem{jing2020self}
Jing, L., Chen, Y., Zhang, L., He, M., Tian, Y.: Self-supervised modal and view
  invariant feature learning. arXiv preprint arXiv:2005.14169  (2020)

\bibitem{xue2023ulip}
Xue, L., Gao, M., Xing, C., Mart{\'\i}n-Mart{\'\i}n, R., Wu, J., Xiong, C., Xu,
  R., Niebles, J.C., Savarese, S.: Ulip: Learning a unified representation of
  language, images, and point clouds for 3d understanding. In: CVPR (2023)

\bibitem{xue2023ulip2}
Xue, L., Yu, N., Zhang, S., Li, J., Mart{\'\i}n-Mart{\'\i}n, R., Wu, J., Xiong,
  C., Xu, R., Niebles, J.C., Savarese, S.: Ulip-2: Towards scalable multimodal
  pre-training for 3d understanding. arXiv preprint arXiv:2305.08275  (2023)

\bibitem{sun2018fishnet}
Sun, S., Pang, J., Shi, J., Yi, S., Ouyang, W.: Fishnet: A versatile backbone
  for image, region, and pixel level prediction. In: NeurIPS (2018)

\bibitem{wang2021unsupervised}
Wang, H., Liu, Q., Yue, X., Lasenby, J., Kusner, M.J.: Unsupervised point cloud
  pre-training via occlusion completion. In: ICCV (2021)

\bibitem{yin2022proposalcontrast}
Yin, J., Zhou, D., Zhang, L., Fang, J., Xu, C.Z., Shen, J., Wang, W.:
  Proposalcontrast: Unsupervised pre-training for lidar-based 3d object
  detection. In: ECCV (2022)

\bibitem{sauder2019self}
Sauder, J., Sievers, B.: Self-supervised deep learning on point clouds by
  reconstructing space. In: NeurIPS (2019)

\bibitem{achlioptas2018learning}
Achlioptas, P., Diamanti, O., Mitliagkas, I., Guibas, L.: Learning
  representations and generative models for 3d point clouds. In: ICML (2018)

\bibitem{li2018so}
Li, J., Chen, B.M., Lee, G.H.: So-net: Self-organizing network for point cloud
  analysis. In: CVPR (2018)

\bibitem{yang2018foldingnet}
Yang, Y., Feng, C., Shen, Y., Tian, D.: Foldingnet: Point cloud auto-encoder
  via deep grid deformation. In: CVPR (2018)

\bibitem{rao2020global}
Rao, Y., Lu, J., Zhou, J.: Global-local bidirectional reasoning for
  unsupervised representation learning of 3d point clouds. In: CVPR (2020)

\bibitem{eckart2021self}
Eckart, B., Yuan, W., Liu, C., Kautz, J.: Self-supervised learning on 3d point
  clouds by learning discrete generative models. In: CVPR. pp. 8248--8257
  (2021)

\bibitem{yan2022iae}
Yan, S., Yang, Z., Li, H., Guan, L., Kang, H., Hua, G., Huang, Q.: Iae:
  Implicit autoencoder for point cloud self-supervised representation learning.
  arXiv preprint arXiv:2201.00785  (2022)

\bibitem{feng2023clustering}
Feng, T., Wang, W., Wang, X., Yang, Y., Zheng, Q.: Clustering based point cloud
  representation learning for 3d analysis. In: ICCV (2023)

\bibitem{min2022voxel}
Min, C., Zhao, D., Xiao, L., Nie, Y., Dai, B.: Voxel-mae: Masked autoencoders
  for pre-training large-scale point clouds. arXiv preprint arXiv:2206.09900
  (2022)

\bibitem{rolfe2016discrete}
Rolfe, J.T.: Discrete variational autoencoders. arXiv preprint arXiv:1609.02200
   (2016)

\bibitem{long2023pointclustering}
Long, F., Yao, T., Qiu, Z., Li, L., Mei, T.: Pointclustering: Unsupervised
  point cloud pre-training using transformation invariance in clustering. In:
  CVPR (2023)

\bibitem{chen20224dcontrast}
Chen, Y., Nie{\ss}ner, M., Dai, A.: 4dcontrast: Contrastive learning with
  dynamic correspondences for 3d scene understanding. In: ECCV (2022)

\bibitem{liu2019point2sequence}
Liu, X., Han, Z., Liu, Y.S., Zwicker, M.: Point2sequence: Learning the shape
  representation of 3d point clouds with an attention-based sequence to
  sequence network. In: AAAI (2019)

\bibitem{tang2020searching}
Tang, H., Liu, Z., Zhao, S., Lin, Y., Lin, J., Wang, H., Han, S.: Searching
  efficient 3d architectures with sparse point-voxel convolution. In: ECCV
  (2020)

\bibitem{chen2022focal}
Chen, Y., Li, Y., Zhang, X., Sun, J., Jia, J.: Focal sparse convolutional
  networks for 3d object detection. In: CVPR (2022)

\bibitem{feng2024lsk3dnet}
Feng, T., Wang, W., Ma, F., Yang, Y.: Lsk3dnet: Towards effective and efficient
  3d perception with large sparse kernels. In: CVPR (2024)

\bibitem{yvette2011noether}
Yvette, K.S.: The noether theorems. invariance and conservation laws in the
  twentieth century. Translated by Bertram E. Schwarzbach. NY: Springer  (2011)

\bibitem{kong2023understanding}
Kong, X., Zhang, X.: Understanding masked image modeling via learning occlusion
  invariant feature. In: CVPR (2023)

\bibitem{dangovski2021equivariant}
Dangovski, R., Jing, L., Loh, C., Han, S., Srivastava, A., Cheung, B., Agrawal,
  P., Soljacic, M.: Equivariant self-supervised learning: Encouraging
  equivariance in representations. In: ICLR (2021)

\bibitem{chen2020simple}
Chen, T., Kornblith, S., Norouzi, M., Hinton, G.: A simple framework for
  contrastive learning of visual representations. In: ICML (2020)

\bibitem{wang2021exploring}
Wang, W., Zhou, T., Yu, F., Dai, J., Konukoglu, E., Van~Gool, L.: Exploring
  cross-image pixel contrast for semantic segmentation. In: ICCV (2021)

\bibitem{chang2015shapenet}
Chang, A.X., Funkhouser, T., Guibas, L., Hanrahan, P., Huang, Q., Li, Z.,
  Savarese, S., Savva, M., Song, S., Su, H., et~al.: Shapenet: An
  information-rich 3d model repository. arXiv preprint arXiv:1512.03012  (2015)

\bibitem{hackel2017semantic3d}
Hackel, T., Savinov, N., Ladicky, L., Wegner, J.D., Schindler, K., Pollefeys,
  M.: Semantic3d. net: A new large-scale point cloud classification benchmark.
  arXiv preprint arXiv:1704.03847  (2017)

\bibitem{loshchilov2017decoupled}
Loshchilov, I., Hutter, F.: Decoupled weight decay regularization. arXiv
  preprint arXiv:1711.05101  (2017)

\bibitem{loshchilovstochastic}
Loshchilov, I., Hutter, F.: Stochastic gradient descent with warm restarts. In:
  ICLR (2016)

\bibitem{phan2018dgcnn}
Phan, A.V., Le~Nguyen, M., Nguyen, Y.L.H., Bui, L.T.: Dgcnn: A convolutional
  neural network over large-scale labeled graphs. Neural Networks
  \textbf{108},  533--543 (2018)

\bibitem{liu2019relation}
Liu, Y., Fan, B., Xiang, S., Pan, C.: Relation-shape convolutional neural
  network for point cloud analysis. In: CVPR (2019)

\bibitem{zheng2023point}
Zheng, X., Huang, X., Mei, G., Hou, Y., Lyu, Z., Dai, B., Ouyang, W., Gong, Y.:
  Point cloud pre-training with diffusion models. arXiv preprint
  arXiv:2311.14960  (2023)

\bibitem{qi2019deep}
Qi, C.R., Litany, O., He, K., Guibas, L.J.: Deep hough voting for 3d object
  detection in point clouds. In: ICCV (2019)

\bibitem{misra2021end}
Misra, I., Girdhar, R., Joulin, A.: An end-to-end transformer model for 3d
  object detection. In: ICCV (2021)

\bibitem{wang2023take}
Wang, Z., Yu, X., Rao, Y., Zhou, J., Lu, J.: Take-a-photo: 3d-to-2d generative
  pre-training of point cloud models. In: ICCV (2023)

\bibitem{huang2021spatio}
Huang, S., Xie, Y., Zhu, S.C., Zhu, Y.: Spatio-temporal self-supervised
  representation learning for 3d point clouds. In: ICCV (2021)

\bibitem{mo2019partnet}
Mo, K., Zhu, S., Chang, A.X., Yi, L., Tripathi, S., Guibas, L.J., Su, H.:
  Partnet: A large-scale benchmark for fine-grained and hierarchical part-level
  3d object understanding. In: CVPR (2019)

\bibitem{song2015sun}
Song, S., Lichtenberg, S.P., Xiao, J.: Sun rgb-d: A rgb-d scene understanding
  benchmark suite. In: CVPR (2015)

\bibitem{boudjoghra20243d}
Boudjoghra, M.E.A., Al~Khatib, S., Lahoud, J., Cholakkal, H., Anwer, R., Khan,
  S.H., Shahbaz~Khan, F.: 3d indoor instance segmentation in an open-world
  (2023)

\bibitem{cen2022open}
Cen, J., Yun, P., Zhang, S., Cai, J., Luan, D., Tang, M., Liu, M., Yu~Wang, M.:
  Open-world semantic segmentation for lidar point clouds. In: ECCV (2022)

\end{thebibliography}

\newpage

\appendix
\setcounter{table}{0}
\setcounter{figure}{0}
\setcounter{footnote}{0}
\renewcommand{\thetable}{A\arabic{table}}
\renewcommand{\thefigure}{A\arabic{figure}}
\renewcommand{\thepage}{A\arabic{page}}

\begin{center}
    \Large{\textbf{Supplementary Material}}
\end{center}

\definecolor{ggray}{RGB}{127,127,127}
\definecolor{mygreen}{RGB}{93,174,86}
\definecolor{mypurple}{RGB}{123,104,238}

In this supplementary material, we provide the following sections to enhance comprehension of the main paper. The experiment setting is elaborated in \S\ref{appendix1}. Subsequently, \S\ref{appendix3} presents details of the network structure for MH-V in the context of 3D object detection and semantic segmentation, as well as the point-point contrastive loss (PPC). Finally, the limitations and social impact are discussed in \S\ref{appendix5}.

\section{Experiment Setting} 
\label{appendix1}

\subsection{License of Assets}

PointGPT~\cite{chen2023pointgpt}\footnote{\url{https://github.com/CGuangyan-BIT/PointGPT/}} has been released under an MIT license. VoteNet~\cite{qi2019deep}\footnote{\url{https://github.com/facebookresearch/votenet/}} has also been implemented under an MIT license. The experiments involve publicly available datasets that have been widely applied for 3D research. ModelNet40~\cite{wu20153d}\footnote{\url{https://modelnet.cs.princeton.edu/}}, ShapeNet~\cite{chang2015shapenet}\footnote{\url{https://shapenet.org//}}, ShapeNetPart~\cite{yi2016scalable}\footnote{\url{https://cs.stanford.edu/~ericyi/project_page/part_annotation/}}, and S3DIS~\cite{armeni20163d}\footnote{\url{http://buildingparser.stanford.edu/dataset.html}} have custom licenses that only allow academic use. Synthia4D~\cite{ros2016synthia}\footnote{\url{https://synthia-dataset.net/}} is published under the Creative Commons Attribution-NonCommercial-ShareAlike 3.0 License. ScanObjectNN dataset~\cite{uy2019revisiting}\footnote{\url{https://hkust-vgd.github.io/scanobjectnn/}} is released under the `ScanObjectNN Terms of Use'. ScanNetv2~\cite{dai2017scannet} is released under the `ScanNet Terms of Use'\footnote{\url{https://kaldir.vc.in.tum.de/scannet/ScanNet_TOS.pdf}}. We use the SemanticKITTI dataset~\cite{behley2019semantickitti} under the permission of its creators and authors by registering at the competition website\footnote{\url{https://codalab.lisn.upsaclay.fr/competitions/6280}}. The unlabeled hybrid dataset (UHD) and labeled hybrid dataset (LHD) are built upon ModelNet40~\cite{wu20153d}, PartNet~\cite{mo2019partnet}, ShapeNet~\cite{chang2015shapenet}, S3DIS~\cite{armeni20163d}, ScanObjectNN~\cite{uy2019revisiting}, SUN RGB-D~\cite{song2015sun}, and Semantic3D~\cite{hackel2017semantic3d}\footnote{\url{https://github.com/CGuangyan-BIT/PointGPT}}, following the licenses of each dataset.

\subsection{Experimental Environment}

Software and hardware environment:
\begin{itemize}[leftmargin=*]
  \setlength{\itemsep}{0pt}
  \setlength{\parsep}{-2pt}
  \setlength{\parskip}{-0pt}
  \setlength{\leftmargin}{-10pt}
  \item CUDA version: 10.1
  \item cuDNN version: 7.6.3
  \item PyTorch version: 1.7.0
  \item GPU: Nvidia Tesla V100, 32GB VRAM
  \item CPU: Intel(R) Xeon(R) Gold 6132 CPU @ 2.60GHz
\end{itemize}

\begin{figure*}[t]
  \centering
      \includegraphics[width=1.0 \linewidth]{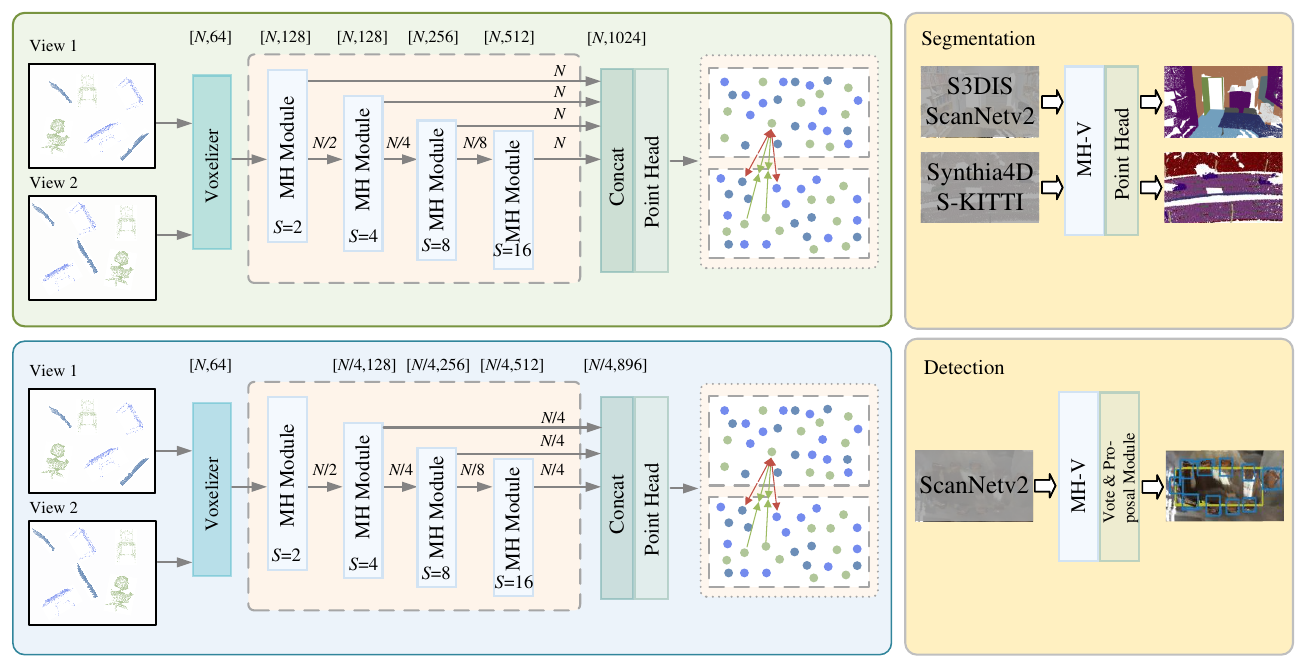}
      \put(-290,95){\scalebox{.80}{(a) Pre-Training for Semantic Segmentation Task}}
      \put(-102,105){\scalebox{.80}{(b) Downstream Semantic}}
      \put(-90,95){\scalebox{.80}{Segmentation Task}}
      \put(-290,7){\scalebox{.80}{(c) Pre-Training for Object Detection Task}}
      \put(-102,17){\scalebox{.80}{(d) Downstream Object}}
      \put(-90,7){\scalebox{.80}{Detection Task}}
      \put(-142,98){\scalebox{.80}{Eq. ({\color{red}1})}}
      \put(-142,11){\scalebox{.80}{Eq. ({\color{red}1})}}
      \put(-264,128){\scalebox{.80}{$x_{2}$}}
      \put(-264,41){\scalebox{.80}{$x_{2}$}}
      \put(-245,128){\scalebox{.80}{$x_{4}$}}
      \put(-245,41){\scalebox{.80}{$x_{4}$}}
      \put(-224,128){\scalebox{.80}{$x_{8}$}}
      \put(-224,41){\scalebox{.80}{$x_{8}$}}
      \put(-204,128){\scalebox{.65}{$x_{16}^h$}}
      \put(-204,41){\scalebox{.65}{$x_{16}^h$}}
      \put(-262,157){\scalebox{.65}{$x_{2}^h$}}
      \put(-242,151){\scalebox{.65}{$x_{4}^h$}}
      \put(-222,145){\scalebox{.65}{$x_{8}^h$}}
      \put(-222,58){\scalebox{.65}{$x_{8}^h$}}
      \put(-242,65){\scalebox{.65}{$x_{4}$}}
\captionsetup{font=small}
\caption{\small Illustration for transferring from shape data to scene-level downstream tasks, \ie, Shape2Scene (\S\!~\ref{appendix3}). (a) shows the pre-training of MH-V backbone for semantic segmentation task; (b) shows the downstream semantic segmentation tasks; (c) shows the pre-training of MH-V backbone for object detection task; and (d) shows the downstream object detection tasks.  (Best viewed with zoom-in.)}
\label{fig:MHV}
\end{figure*}

\section{Voxel-based MH Backbone (MH-V)} 
\label{appendix3}

Figs.\!~\ref{fig:MHV} (a-d) demonstrate the  pre-training and fine-tuning of MH-V for scene-level downstream tasks, encompassing 3D semantic segmentation and object detection tasks. The pre-training data is obtained by S2SS. Eq. ({\color{red}1}) represents PPC. The details of S2SS and PPC can be found in the main paper.

\subsection{MH-V for 3D Semantic Segmentation}

Fig.\!~\ref{fig:MHV} (a) visually demonstrates the pre-training process of MH-V backbone for the semantic segmentation task. It is constructed atop four MH modules (as referenced in Fig.\!~{\color{red}2} (d) of the main paper). These modules are configured to operate at scales of 2, 4, 8, and 16, respectively. Specifically, within the MH module designated for scale $S$, its input derives from the output $x_{s'}$ of the former MH module (with a scale of $S'$). Additionally, $x_{s}$ transitions into the input for the subsequent MH module. Similar to MH-P backbone, high-resolution mapping is implemented within each MH module, seamlessly integrating all ${x_s^h}$ as the input for the point head.$_{\!}$ This innovative architecture$_{\!}$ empowers$_{\!}$ MH-V$_{\!}$ with direct access$_{\!}$ to$_{\!}$ high-resolution$_{\!}$ data$_{\!}$ and$_{\!}$ integrated features$_{\!}$ spanning various scales. Subsequent to the pre-training phase, we apply MH-V backbone to the downstream semantic segmentation task, as illustrated  Fig.\!~\ref{fig:MHV} (b).

\subsection{MH-V for 3D Object Detection}

Figs.\!~\ref{fig:MHV} (c-d) show the MH-V backbone for pre-training and the downstream object detection task, respectively. The backbone is built on the top of four MH modules (see Fig.\!~{\color{red}2} (d) in the main paper). The modules are operated at scales of 2, 4, 8, and 16, respectively. The MH module used here shares the same structure as the one employed in the semantic segmentation task. The difference lies in that the integration of ${x_{4}}$, ${x_{8}^h}$, and ${x_{16}^h}$ is taken as the input of point head. The high-resolution features, at a scale of 4, are mapped from features on scales of 8 and 16. This is because the 3D object detection task focuses on region-level features. Moreover, 4,096 points at a scale of 4 are randomly selected and then applied to generate point pairs for Eq. ({\color{red}1}). After the pre-training phase, we apply MH-V as the backbone and use the vote \& proposal module~\cite{qi2019deep} to detect objects, as illustrated in Fig.\!~\ref{fig:MHV} (d).

\subsection{Pseudo-Code for Point-Point Contrastive Loss}
Here, we present a PyTorch-style pseudo-code for point-point contrastive loss (PPC).

\begin{algorithm}[H]
\caption{\small Pseudo-code of PPC.}
\label{alg:code}
\definecolor{codegreen}{rgb}{0,0.6,0}
\definecolor{codegray}{rgb}{0.5,0.5,0.5}
\definecolor{codepurple}{rgb}{0.58,0,0.82}
\definecolor{backcolour}{rgb}{0.95,0.95,0.92}
\lstdefinestyle{mystyle}{
  backgroundcolor=\color{backcolour}, commentstyle=\color{codegreen},
  keywordstyle=\color{magenta},
  numberstyle=\tiny\color{codegray},
  stringstyle=\color{codepurple},
  basicstyle=\ttfamily\fontsize{6.9}{11}\linespread{0.8}\selectfont,
  breakatwhitespace=false,         
  breaklines=true,                 
  captionpos=b,                    
  keepspaces=true,                 
  numbers=left,                    
  numbersep=3pt,                  
  showspaces=false,                
  showstringspaces=false,
  showtabs=false,                  
  tabsize=2,
}
\lstset{style=mystyle}
\begin{lstlisting}[language=python]
# Z_1, Z_2: features for matched points between view 1 and view 2: (M x 2048, C)
# t: temperature
# Ns: subsampling size for point features.

# mark each point
mark = torch.arange(M).repeat_interleave(2048)
mark = mark.view(-1,1)
pairs = torch.eq(mark,mark.T)
# get positive pairs 
pos_pair = torch.where(pairs==True)
# get subsampling indexes
pos_pair_num = pos_pair[0].shape[0]
inds = random.sample(range(pos_pair_num), Ns)
# get subsample point features
Z_1 = Z_1[pos_pair[0][inds],:] 
Z_2 = Z_2[pos_pair[1][inds],:]
sim = torch.mm(Z_1,Z_2.T) # Ns x Ns
labels = torch.arange(Ns) 
loss = CrossEntropyLoss(sim/t,labels)
\end{lstlisting}
\end{algorithm}

\section{Limitation and Social Impact}
\label{appendix5}

\subsection{Limitation}

We have extensively researched three classical shape-level downstream tasks and five scene-level benchmarks: S3DIS~\cite{armeni20163d}, Synthia4D\!~\cite{ros2016synthia}, ScanNet v2\!~\cite{dai2017scannet}, and  Semantic-
KITTI\!~\cite{behley2019semantickitti}. Our research has narrowed the gap between 3D shape and scene-level datasets. However, there are remaining challenges, including expanding S2S's applications to broader 3D scenarios, encompassing more challenging open-world scenes. Open world of point clouds (see \cite{boudjoghra20243d,cen2022open}) is a new downstream task, S2S may be helpful for it. We will address this challenges and conduct research on open-world 3D perception tasks, such as on ScanNet v2\!~\cite{boudjoghra20243d} and SemanticKITTI\!~\cite{cen2022open}, in our future work.

\subsection{Social Impact}

With the potential emergence of large-scale shape-level datasets, exploring research using a pretext task grounded in these datasets gains increased significance. Within real-world contexts, endeavors concentrating on scene-level perception are closely linked to augmented reality, urban planning, and autonomous vehicles. This research contributes to bridging the divide between shape-level and scene-level datasets. It holds promise for advancing various facets of 3D technologies, spanning from 3D self-supervised learning to scene comprehension and beyond.

\end{document}